\documentclass{worldbench}

\definecolor{w_blue}{RGB}{52,204,204}
\definecolor{w_yellow}{RGB}{255,192,0}
\definecolor{w_red}{RGB}{192,0,0}

\usepackage{algorithmic}
\usepackage{algorithm}
\usepackage{amsmath}
\usepackage{amssymb}
\usepackage{amsthm}

\usepackage{booktabs}
\usepackage{adjustbox}

\usepackage{caption}
\usepackage{color}
\usepackage{colortbl}

\usepackage{enumitem}

\usepackage{graphicx}
\usepackage{booktabs}
\usepackage{array}
\usepackage{xltabular}
\usepackage{hhline}
\usepackage{longtable}

\usepackage{makecell}
\usepackage{mathtools}
\usepackage{microtype}
\usepackage{multicol}
\usepackage{multirow}

\usepackage{pifont}

\usepackage{subcaption}

\usepackage{tabularx}
\usepackage{tikz}
\usetikzlibrary{arrows.meta, positioning}
\usepackage{stfloats}

\usepackage{wrapfig}

\usepackage{xspace}

\usepackage{titletoc}

\definecolor{crKITTI}{RGB}{0, 102, 102}
\definecolor{crNYU}{RGB}{83, 22, 136}
\definecolor{crCarla}{RGB}{153, 153, 0}
\definecolor{crSemanticKITTI}{RGB}{231, 76, 60}
\definecolor{crnuScenes}{RGB}{89, 74, 235}
\definecolor{crWaymo}{RGB}{106, 228, 163}
\definecolor{crSeeingThroughFog}{RGB}{96, 96, 96}
\definecolor{vKITTI}{RGB}{202, 200, 229}
\definecolor{crArgoverse}{RGB}{255, 128, 0}
\definecolor{crLyft-Level5}{RGB}{235, 11, 140}
\definecolor{crnuPlan}{RGB}{52, 204, 204}
\definecolor{crPandaSet}{RGB}{32, 32, 32}
\definecolor{crOpenCOOD}{RGB}{0, 0, 255}
\definecolor{crKITTI360}{RGB}{165, 105, 189}
\definecolor{crCarlaSC}{RGB}{0, 176, 240}
\definecolor{crRobo}{RGB}{69, 82, 104}
\definecolor{crOpenOcc}{RGB}{255, 0, 255}
\definecolor{crOcc3D-nuScenes}{RGB}{89, 74, 235}
\definecolor{crOpenDV}{RGB}{255, 51, 51}
\definecolor{crSSCBench}{RGB}{83, 22, 136}
\definecolor{crNAVSIM}{RGB}{153, 0, 76}
\definecolor{crDrivingDojo}{RGB}{152, 77, 16}
\definecolor{EUVS}{RGB}{187, 155, 209}
\definecolor{crOmniDrive}{RGB}{155, 200, 26}
\definecolor{crPi3DET}{RGB}{239, 99, 75}
\definecolor{crPrivate}{RGB}{90, 90, 90}
\definecolor{cotink}{HTML}{17222F}
\definecolor{cotmut}{HTML}{5F6F81}
\definecolor{cotline}{HTML}{D9E1EA}
\definecolor{cotblue}{HTML}{1E40AF}
\definecolor{cotred}{HTML}{C22A2A}
\definecolor{cotredln}{HTML}{EFC7C3}
\definecolor{cotredbg}{HTML}{FDF4F3}
\definecolor{cotgrn}{HTML}{0B7A55}
\definecolor{cotgrnln}{HTML}{C9E8DA}
\definecolor{cotgrnbg}{HTML}{EDF8F3}
\definecolor{cotcard}{HTML}{FBFCFE}

\definecolor{crNuplanOcc}{RGB}{214, 152, 60}
\definecolor{crLidarOpen}{RGB}{80, 160, 175}
\definecolor{crOccInteract}{RGB}{175, 120, 195}

\definecolor{crWorldLens}{RGB}{60, 130, 200}

\newcolumntype{C}[1]{>{\centering\arraybackslash}m{#1}}
\newcolumntype{L}[1]{>{\raggedright\arraybackslash}p{#1}}
\newcolumntype{Y}{>{\raggedright\arraybackslash}X}
\newcolumntype{Z}{>{\centering\arraybackslash}X}
\definecolor{VLAbg}{HTML}{FFFBE8}
\definecolor{WAMbg}{HTML}{EEF6FF}

\newtcolorbox{wbcard}[1]{%
  enhanced, breakable,
  colback=wblue!4, colframe=wblue,
  boxrule=0.7pt, arc=4pt,
  left=9pt, right=9pt, top=6pt, bottom=6pt,
  before skip=8pt, after skip=8pt,
  fonttitle=\sffamily\bfseries\large, coltitle=white,
  colbacktitle=wblue, title={#1}
}
\newtcolorbox{wbtakeaway}{%
  enhanced, breakable,
  colback=wblue!5, colframe=wblue,
  boxrule=0.5pt, arc=3pt,
  left=9pt, right=9pt, top=5pt, bottom=5pt,
  before skip=7pt, after skip=7pt,
  borderline west={2.5pt}{0pt}{wblue},
  fonttitle=\sffamily\bfseries, coltitle=wfg,
  title={Takeaway}
}
\definecolor{layerReal}{HTML}{C0392B}
\definecolor{layerEgo}{HTML}{8E44AD}
\definecolor{layerUMI}{HTML}{1F9E9E}
\definecolor{layerSim}{HTML}{2E7D32}
\definecolor{layerGen}{HTML}{34495E}
\definecolor{layerApp}{HTML}{F5B027}
\newtcbox{\layerbadge}[1][gray]{%
  on line, arc=2pt, colback=#1, colframe=#1, boxrule=0pt,
  left=3pt, right=3pt, top=1pt, bottom=1pt,
  fontupper=\scriptsize\sffamily\bfseries\color{white}
}

\DeclareRobustCommand{\DataLegendItem}[2]{%
  \begingroup
    \edef\SavedCaptionFont{\the\font}%
    \raisebox{-0.18\height}{%
      \resizebox{!}{1.15em}{{#1}}%
    }%
    \SavedCaptionFont\nobreakspace #2%
  \endgroup
}

\titlecontents{section}
  [0.45em]
  {\normalsize\sffamily\bfseries\vspace{4pt}}
  {\textcolor{wblue}{\contentslabel{1.65em}}}
  {}
  {\titlerule*[0.45pc]{.}\contentspage}
\titlecontents{subsection}
  [2.1em]
  {\normalsize\sffamily\vspace{1pt}}
  {\textcolor{wblue}{\contentslabel{2.35em}}}
  {}
  {\titlerule*[0.45pc]{.}\contentspage}

\title{Large Multimodal Models for Direct Robot Control: GPT-6 as a Beginning}

\author[1, *]{Youwei Zhou}
\author[2, *]{Bohan Hou}
\author[3, *]{Yankai Fu}
\author[3, *]{Yaoxu Lyu}
\author[5]{Yifan Ye}
\author[2]{Shilin Shan}
\author[5]{Chen Guo}
\author[3]{\protect\\ Jiaming Liu}
\author[4]{Ran He}
\author[1]{Caifeng Shan}
\author[3]{Shanghang Zhang}
\author[2, \dagger]{Jianfei Yang}
\author[1, \dagger]{Zhihe Lu}

\affiliation[1]{Nanjing University}
\affiliation[2]{Nanyang Technological University}
\affiliation[3]{Peking University}
\addtolist[4]{Institute of Automation of the Chinese Academy of Sciences}{\affiliationlist}{\affiliationformat}{,\protect\\}
\affiliation[5]{Independent Researcher}

\affiliationlogos{%
    \includegraphics[height=0.8cm]{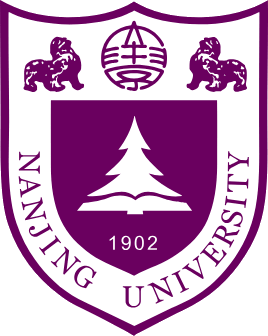}
    \hspace{0.1cm}
    \includegraphics[height=0.8cm]{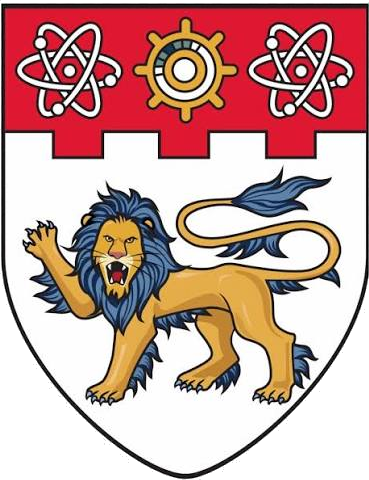}
    \hspace{0.1cm}
    \includegraphics[height=0.8cm]{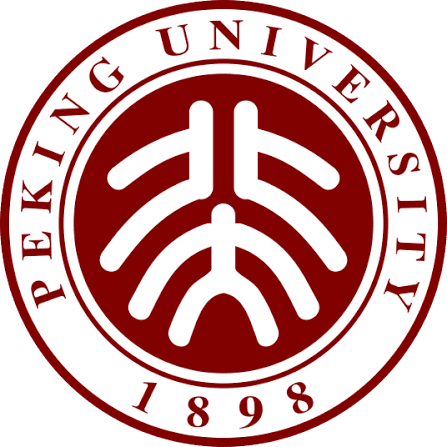}
    \hspace{0.1cm}
    \includegraphics[height=0.8cm]{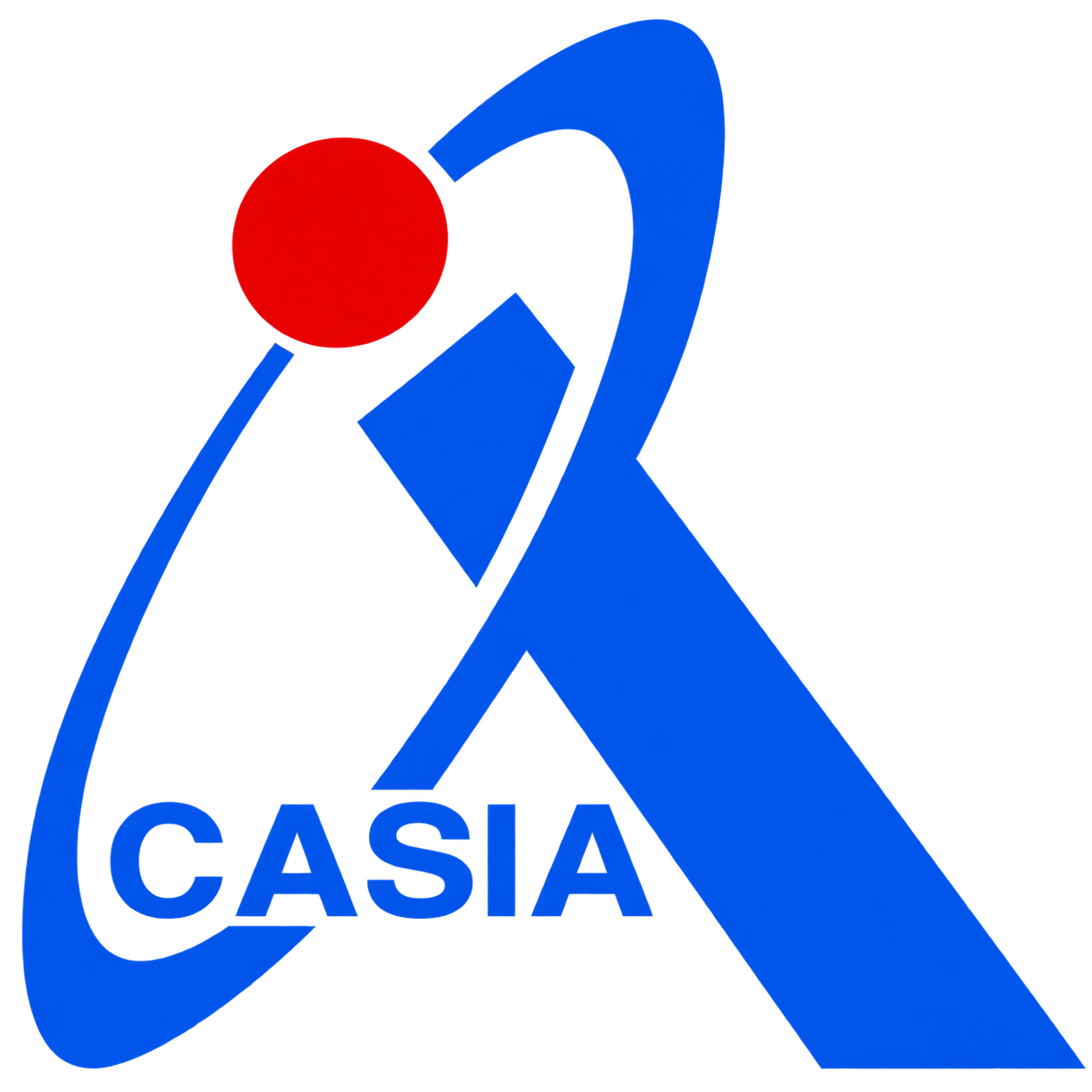}
}

\contribution[*]{Equal contribution}
\contribution[\dagger]{Corresponding authors}

\project{\url{https://6uooon9.github.io/GPT-GEAR}}
\github{\url{https://github.com/6UOOON9/GPT-GEAR}}
\date{\today}

\abstract{
\begin{abstract}~Recent advances in visuospatial understanding and agentic reasoning open the possibility of using general-purpose multimodal models for direct robotic manipulation without specialized action policies; however, their embodied capabilities and limitations remain poorly understood. This paper presents a systematic evaluation of GPT-6 Astra for closed-loop robotic manipulation, built on a lightweight agent execution framework. We first probe three fundamental embodied capabilities—depth perception, embodied in-context learning, and tactile understanding—and demonstrate that the model can exploit depth information, acquire simple manipulation strategies from contextual demonstrations, and use tactile feedback to correct its actions. We then evaluate GPT-6 Astra across five established manipulation benchmarks along six capability dimensions. Despite competitive performance on several benchmarks—26.92\% success on RoboDojo and 76.0\% on RoboTwin 2.0—our analysis reveals a pronounced gap between semantic reasoning and physical execution, particularly in consequence-aware action planning, precision, and complex motion coordination. To further examine these limitations, we introduce LIBERO-GeoMotion, a new benchmark of geometry-constrained tasks that require precise pose alignment, coordinated translation and rotation, and anticipation of physical action consequences. GPT-6 Astra achieves only 11.1\% zero-shot success, exposing the difficulty of translating spatial understanding into physically feasible motion. Real-world experiments corroborate these findings: zero-shot execution is feasible for simple tasks, yet persistent deficits remain in motion feasibility, precise control, and physical safety, while substantial inference latency further hinders practical deployment. 
Overall, our results characterize the emerging embodied capabilities of general-purpose multimodal models and delineate the key challenges toward safe, efficient, and reliable robotic manipulation.
\end{abstract}
}

\metadata[Keywords]{Robotic manipulation, Embodied agent, GPT-6 Astra, \\ Embodied systematic analysis, Large multimodal models}

\begin{document}

\maketitle

\section{Introduction}

Robot manipulation requires a tightly coupled set of capabilities: understanding language and visual scenes, reasoning about spatial relationships, generating physically grounded actions, and adapting to environmental feedback~\cite{bai2025unified, xu2024robotics}. 
Although general-purpose LLMs and VLMs excel at semantic understanding and high-level planning~\cite{ahn2022saycan, driess2023palme}, their limited spatial grounding~\cite{chen2024spatialvlm} and closed-loop control preclude reliable action generation~\cite{huang2023innermonologue, liang2023code, li2026rynnbrain, dang2026rynnbrain}. 
LLM/VLM-based coding agents have likewise been explored for robotic manipulation~\cite{liang2023code}; however, they specify only high-level action logic, whereas precise manipulation points and target poses are still delegated to hand-engineered geometric computations and control primitives, which limits their flexibility.

In contrast, specialized action models, such as Vision-Language-Action (VLA)~\cite{brohan2023rt1, brohan2023rt2, kim2025openvla, black2025pi0, ghosh2024octo,bjorck2025gr00t,ye2026self,ye2026token} models and World Action Models (WAMs)~\cite{cen2025worldvla, wang2026openwam, hou2026world}, directly generate actions, but remain constrained by their embodied training distributions~\cite{ye2026data}, which limits their generalization to unseen tasks and environments~\cite{fei25libero-plus}. Agentic hybrid architectures~\cite{shi2025hirobot, zhang2026harnessvla, peng2026cortex} aim to combine the strengths of both families, using general models for planning and specialized action models for execution. However, these systems still rely on post-trained action policies, even for relatively simple manipulation tasks, and therefore inherit the same generalization and flexibility bottlenecks.


\begin{figure*}[!t]
    \centering
    \includegraphics[width=\textwidth, trim={0 5mm 0 5mm}]{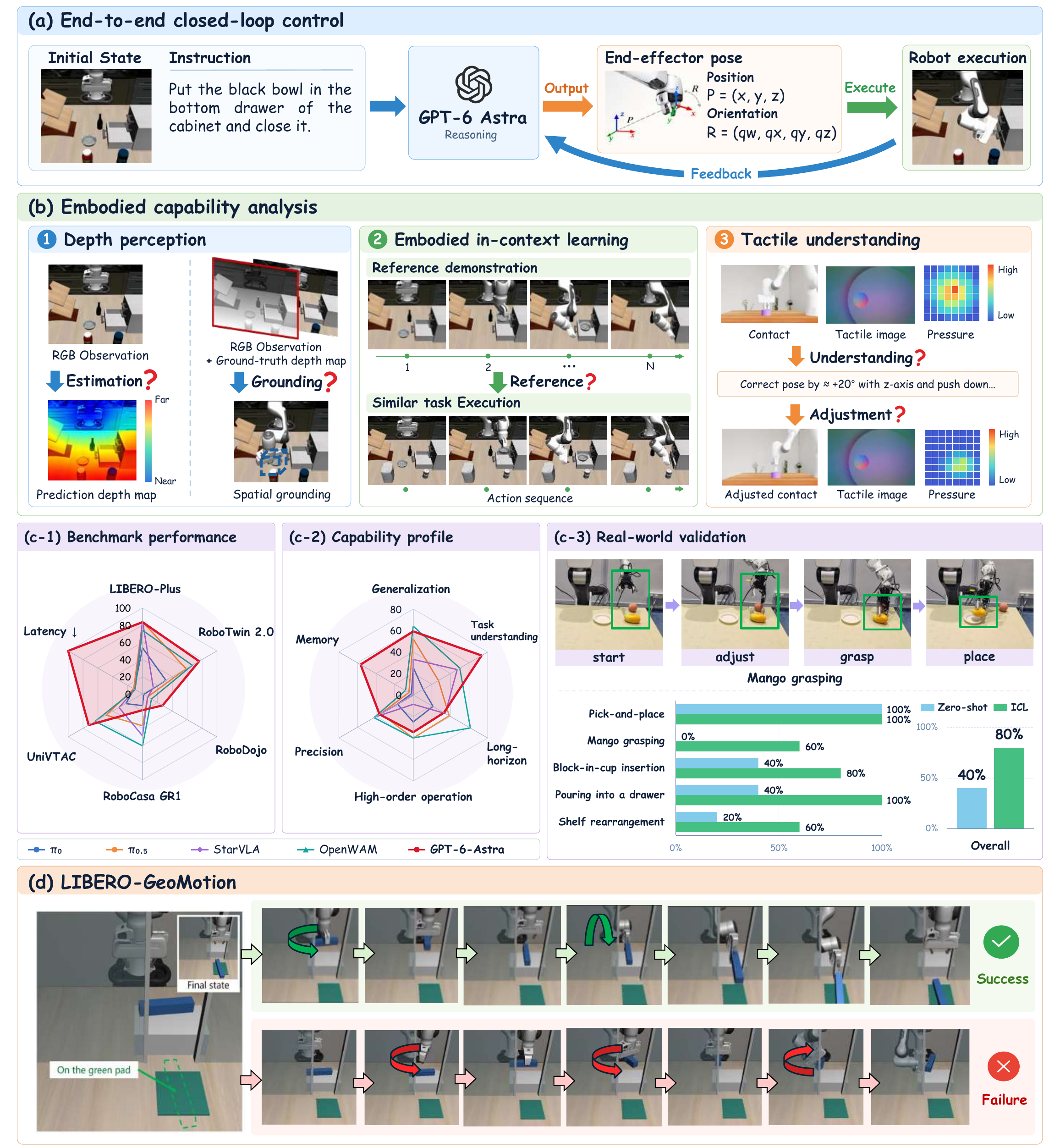}
    \caption{\textbf{Overview of the systematic evaluation of GPT-6 Astra for robotic manipulation.} (a) illustrates the closed-loop control framework. (b) represents the analysis of depth perception, embodied in-context learning, and tactile understanding. (c) summarizes benchmark performance, embodied capability profile and real-world validation. (d) highlights limitations in rotation orientation control through benchmark LIBERO-GeoMotion.}
    \label{fig:motivation}
    \vspace{-5bp}
\end{figure*}

Recently, GPT-6 Astra~\cite{openai2026astra} has emerged as a promising candidate, with strong visuospatial understanding, 3D reasoning, and agentic reasoning. 
These capabilities suggest that a general-purpose model could directly generate executable actions through spatial reasoning and sustain closed-loop control through agentic reasoning—without the hand-engineered primitives or post-trained action policies required by existing hierarchical architectures. 
Preliminary robotics studies~\cite{su2026gpt_as_policy, zhang2026gpt6_astra_robodojo, cheng26incontext_robot_learning, liu2026roboicl} offer initial evidence for this potential, reporting promising manipulation performance without specialized VLA or WAM policies.
Yet the boundaries of GPT-6 Astra's embodied capabilities remain insufficiently characterized. 
Existing studies focus on benchmark performance and selected failure cases, without establishing which capabilities the model executes reliably and where its limitations emerge across diverse physical demands; in particular, real-world evaluations have not systematically probed precise geometric constraints, complex motion coordination, or safety-critical interactions. 
These gaps raise a fundamental question: \textit{What can GPT-6 Astra reliably achieve in manipulation, and where do its capabilities break down, particularly in the real world?}



To answer this question, we conduct a comprehensive, capability-driven investigation of GPT-6 Astra spanning more than 2,000 trials across simulation and real-world environments. We first introduce GPT-Generalist End-effector Action Reasoning (\textbf{GPT-GEAR}), an agent execution framework that equips GPT-6 Astra with minimally engineered general-purpose tools for direct closed-loop robotic manipulation (Figure~\ref{fig:motivation} (a)). Using GPT-GEAR, we isolate three capabilities that underlie successful manipulation—depth perception~\cite{huang2023voxposer, huang2024rekep}, embodied in-context learning~\cite{cheng26incontext_robot_learning, liu2026roboicl}, and tactile understanding~\cite{li2020tactile, chen2026univtac, lou2026dream, fu2026decal} (Figure~\ref{fig:motivation} (b))—and show that the model can estimate and exploit depth, acquire simple motor skills from in-context demonstrations, and use tactile feedback to correct its actions during execution. We then evaluate GPT-6 Astra on five established benchmarks—LIBERO-Plus~\cite{fei25libero-plus}, RoboTwin 2.0~\cite{chen2025robotwin2}, RoboDojo~\cite{chen2026robodojo}, RoboCasa GR1~\cite{robocasa2024, bjorck2025gr00t}, and UniVTAC~\cite{chen2026univtac} (Figure~\ref{fig:motivation} (c))—providing a broad empirical baseline for general-purpose models in robotic manipulation. Characterizing performance along six capability dimensions reveals an uneven profile: without task-specific policy training, GPT-6 Astra exhibits strong task understanding, promising generalization, and superior memory and attains competitive results on several benchmarks, yet these strengths do not consistently translate into reliable physical execution—particularly in long-horizon manipulation, precision-sensitive tasks, and complex motion coordination—while substantial inference overhead further underscores the gap between general-purpose reasoning and efficient robotic control~\cite{li2026efficient}.

Geometry-constrained manipulation exposes this gap most clearly, as it requires precise pose control, coordinated translation and rotation, and anticipation of physical interactions. To probe these demands, we introduce \textbf{LIBERO-GeoMotion}, a targeted benchmark built on LIBERO~\cite{liu2023libero} assets that emphasizes complex motion execution while minimizing semantic complexity (Figure~\ref{fig:motivation} (d)). On LIBERO-GeoMotion, GPT-6 Astra exhibits substantial difficulties in satisfying geometric and sequential constraints, exposing a critical disconnect between understanding desired spatial configurations and generating physically feasible action sequences. We finally extend the evaluation to real-world zero-shot deployment: GPT-6 Astra reliably completes basic pick-and-place tasks, but its overall zero-shot success rate reaches only 40\%, with some constrained tasks failing entirely, and persistent failures involving unsafe grasp configurations, motion feasibility, and precise alignment reveal substantial challenges in safety-aware planning and reliable physical execution. General-purpose models thus already exhibit meaningful zero-shot manipulation capabilities, but remain far from delivering consistently safe and reliable robotic control.

In summary, our contributions are as follows:
\begin{itemize}
    \item We introduce \textbf{GPT-GEAR}, a lightweight agent execution framework that equips GPT-6 Astra with minimally engineered general-purpose tools, enabling direct closed-loop robotic manipulation \textit{with no specialized action policies}.

    \item We systematically investigate three fundamental embodied capabilities---depth perception, embodied in-context learning, and tactile understanding---revealing how diverse forms of embodied information support action generation and execution.

    \item We conduct comprehensive evaluations across five established manipulation benchmarks and characterize performance along six capability dimensions, establishing a broad empirical baseline and identifying the strengths and limitations of general-purpose models in robotic manipulation.

    \item We introduce \textbf{LIBERO-GeoMotion}, a targeted benchmark for geometry-constrained manipulation that probes precise pose control, coordinated rotational and translational motion, and consequence-aware action planning, revealing critical limitations that conventional pick-and-place benchmarks cannot reveal.

    \item We validate GPT-6 Astra on real robotic hardware, demonstrating its potential for zero-shot execution of simple manipulation tasks while identifying critical limitations in safety-aware planning, motion feasibility, and reliable physical control.
\end{itemize}
\section{Related Work}

\subsection{LLM and VLM-Based system}
Large Language Models (LLMs)~\cite{ahn2022saycan, huang2023innermonologue, liang2023code} and Vision-Language Models (VLMs)~\cite{huang2023voxposer, driess2023palme, huang2024rekep, liu2026guava, chen2026showharness,zhang2025cola,yu2023task,li2023graphadapter,lu2023beyond,lu2025task} support robotic manipulation through semantic reasoning and task planning.
SayCan~\cite{ahn2022saycan} grounds language-based skill selection in learned affordances, while Inner Monologue~\cite{huang2023innermonologue} incorporates environmental feedback into language-based planning.
PaLM-E~\cite{driess2023palme} integrates visual and state observations for embodied reasoning, and Code as Policies~\cite{liang2023code} generates executable programs over perception and control APIs.
VoxPoser~\cite{huang2023voxposer} and ReKep~\cite{huang2024rekep} further ground model outputs through 3D value maps and relational keypoint constraints, respectively.
More recently, Guava~\cite{liu2026guava} enables iterative interaction through semantic manipulation tools, while Show-Harness~\cite{chen2026showharness} translates fine-grained semantic action units into robot movements, supporting closed-loop execution without a separate VLA policy.
Building on these studies, we introduce an agent execution framework with minimally engineered, general-purpose interaction tools for closed-loop control.
This framework supports a comprehensive evaluation of the general model's native manipulation capabilities, revealing its potential and limitations across diverse tasks and environments.

\subsection{VLA and WAM-based system}
Vision-Language-Action (VLA) models~\cite{brohan2023rt1, brohan2023rt2, kim2025openvla, black2025pi0, physicalintelligence2025pi05} connect visual observations and language instructions to robot actions through embodied training.
RT-1~\cite{brohan2023rt1} demonstrates scalable language-conditioned visuomotor learning, while RT-2~\cite{brohan2023rt2} transfers knowledge from vision-language pretraining to robotic control by representing actions as tokens.
OpenVLA~\cite{kim2025openvla} provides an open-source VLA trained on diverse robot demonstrations, and $\pi_0$~\cite{black2025pi0} combines a pretrained VLM with flow matching for continuous action generation.
$\pi_{0.5}$~\cite{physicalintelligence2025pi05} further improves open-world generalization through co-training on heterogeneous robot and non-robot data.
These advances demonstrate that learned action models can inherit semantic knowledge while acquiring increasingly diverse manipulation skills.

World Action Models (WAMs)~\cite{cen2025worldvla, wang2026openwam} further connect action generation with predictive modeling of the environment.
WorldVLA~\cite{cen2025worldvla} jointly models actions and future images within an autoregressive framework, enabling action learning and visual prediction to support each other.
More recently, OpenWAM~\cite{wang2026openwam} systematically studies how video-generative priors, world--action information exchange, and embodied pretraining contribute to manipulation and generalization.
Despite their different architectures, both VLAs and WAMs acquire action-generation capabilities through embodied training, and their transfer to unfamiliar settings remains dependent on data coverage and learned control representations.
Our work evaluates the native manipulation capabilities of a general-purpose model through closed-loop interaction with general-purpose tools, without relying on a separately trained VLA or WAM.

\subsection{Hybrid Architectures}
Hybrid architectures~\cite{shi2025hirobot, peng2026cortex, zhang2026harnessvla} combine the semantic reasoning of general LLMs or VLMs with the execution capabilities of learned action models.
Hi Robot~\cite{shi2025hirobot} uses a high-level VLM to interpret complex instructions and user feedback, and a low-level VLA to execute the resulting commands.
Cortex~\cite{peng2026cortex} further aligns high-level planning with low-level VLA execution through a structured subtask interface and execution-aware training.
Harness VLA~\cite{zhang2026harnessvla} exposes a frozen VLA as a callable primitive for contact-rich manipulation and combines it with analytic primitives and memory-guided planning.
These systems illustrate how reasoning and learned control can complement each other, including settings where the action model remains frozen during deployment.
However, their performance reflects the joint contributions of the reasoning model, action policy and interface, making it difficult to attribute successful manipulation to the general model alone.
Our systematic evaluation removes the separate learned action model, enabling systematic evaluation of the general model's native manipulation capabilities and limitations.

\subsection{GPT-6 Astra Driven Policy}
Recent technical reports~\cite{su2026gpt_as_policy, zhang2026gpt6_astra_robodojo, cheng26incontext_robot_learning,liu2026roboicl,guo2026robodawn} explore GPT-6 Astra as a closed-loop manipulation policy.
GPT-as-Policy~\cite{su2026gpt_as_policy} compares direct control with VLA-assisted execution, while Zhang et al.~\cite{zhang2026gpt6_astra_robodojo} evaluate Astra on RoboDojo, identifying semantic strengths and limitations in precise control.
GPT-Policy~\cite{cheng26incontext_robot_learning} investigates adaptation from demonstrations and interaction feedback, while RoboICL~\cite{liu2026roboicl} combines demonstration context with persistent interaction memory to support in-context robot learning.
RoboDawn~\cite{guo2026robodawn} further explores control through a discrete action interface guided by in-context demonstrations.
These studies demonstrate Astra's embodied potential, but their emphasis on particular benchmarks or settings provides only a partial account of its manipulation capabilities and boundaries.
Furthermore, we provide a more comprehensive capability-level evaluation by jointly examining depth perception, embodied in-context learning and tactile understanding within a common execution framework.
\section{Solution}
\begin{figure*}[!t]
    \centering
    \includegraphics[width=\textwidth]{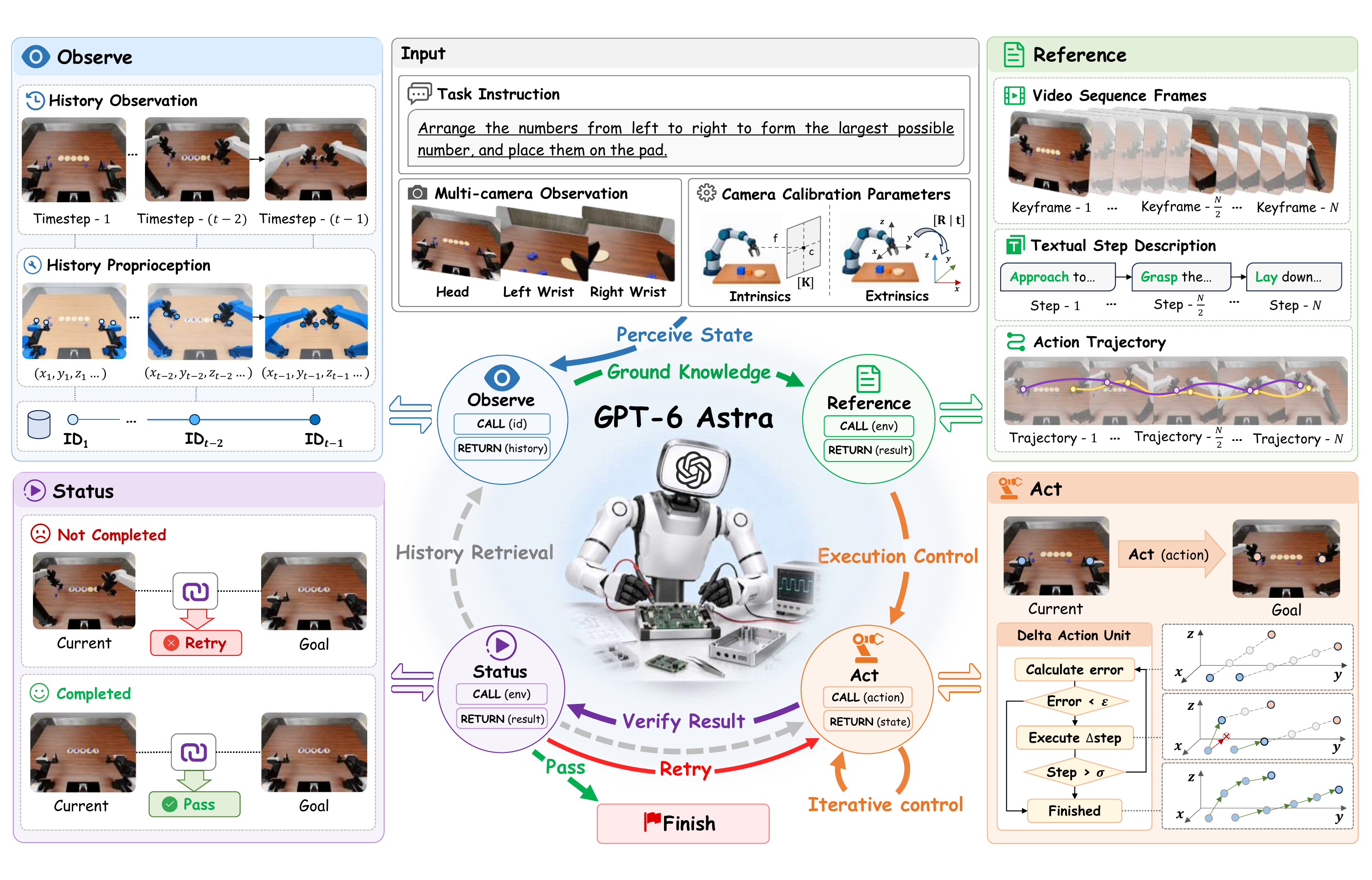}
    \caption{\textbf{GPT-GEAR: a tool-mediated framework for robotic manipulation driven by GPT-6 Astra.} The model generates commands from multimodal inputs and execution feedback. \texttt{Observe} retrieves current or historical observations, \texttt{Reference} supplies optional demonstration context, and \texttt{Act} executes model-generated commands. \texttt{Status} verifies task completion, while \texttt{Finish} terminates the interaction.}
    \label{fig:framework}
\end{figure*}

\subsection{Task Formulation}
We study direct visual control of a robot manipulator as a partially observable, terminal-goal decision problem. 
Given a language instruction $\ell$, the model generates actions from observations and interaction history to achieve the task goal in an environment with hidden physical state $x_t$, where $t$ indexes decision turns. Each observation contains RGB images and proprioception, supplemented by optional inputs indexed by $\mathcal{M}$:
\begin{equation}
    o_t(\mathcal{M}) = \big(\{I_t^c\}_{c\in\mathcal{C}},\,s_t,\,\{v_t^m\}_{m\in\mathcal{M}}\big),
\end{equation}
where $\mathcal{C}$ indexes the available cameras, $I_t^c$ is the RGB image from camera $c$, $s_t$ is the robot proprioceptive state, and $v_t^m$ is an optional input such as camera calibration, depth maps, or tactile feedback.
Varying $\mathcal{M}$ defines the observation conditions, with $\mathcal{M}=\varnothing$ retaining only RGB and proprioception.
At each decision turn, the model selects a tool and generates its arguments:
\begin{equation}
    u(a)_t = (\tau_t,a_t)
    \sim \pi_\theta(\cdot\mid\mathcal{H}_t,\ell),
\end{equation}
where $\tau_t$ is the tool type, $a_t$ denotes its arguments, and $\mathcal{H}_t$ is the ordered history of observations, tool calls, and results. Only \texttt{Act} changes the physical state. In simulation, success is determined by the environment's official predicate $F(x,\ell)$ at termination, independently of the model's completion claim.

\begin{algorithm}[h]
\caption{Closed-loop Execution}
\label{alg:closed_loop}
\begin{algorithmic}[1]
\item[] \textbf{Model judgments:}
\item[] \texttt{ModelDone()}: The model judges the task complete
\item[] \texttt{NeedHistory()}: The model requests historical observations
\item[] \textbf{Procedure:}
\STATE \texttt{Observe()} \COMMENT{Get initial observation}
\STATE \texttt{Reference()} \COMMENT{Load demonstration context}
\WHILE{execution budget remains}
    \STATE \texttt{Act(action)} \COMMENT{Execute and receive feedback}
    \IF{\textbf{not} \texttt{ModelDone()}}
        \IF{\texttt{NeedHistory()}}
            \STATE \texttt{Observe(ID)} \COMMENT{Retrieve a historical record}
        \ENDIF
    \ELSE
        \IF{\texttt{Status()}}
            \STATE \texttt{Finish()}
            \RETURN
        \ENDIF
    \ENDIF
\ENDWHILE
\RETURN \COMMENT{Budget exhausted}
\end{algorithmic}
\end{algorithm}

\subsection{Tool-mediated Agent Architecture}
As shown in Figure~\ref{fig:framework}, GPT-Generalist End-effector Action Reasoning (GPT-GEAR) separates model-level decisions from low-level execution~\cite{liang2023code}. GPT-6 Astra interprets observations and generates commands, while the tool interface provides environmental information and executes the specified actions. Each tool call and its result $z_t$ are appended to the interaction history:
\begin{equation}
    \mathcal{H}_{t+1}
    = \mathcal{H}_t \mathbin{\Vert} \big(u(a)_t,z_t\big),
\end{equation}
where $\Vert$ denotes sequence concatenation. Subsequent decisions are conditioned on this updated history, allowing the model to respond to execution outcomes~\cite{huang2023innermonologue}. This common interaction structure supports capability evaluation by varying sensory evidence and demonstration context. The next subsection describes the tool-mediated workflow.

\subsection{Closed-loop Execution}
We organize five complementary tools into a closed-loop workflow, as illustrated in Figure~\ref{fig:framework}.
\texttt{Observe} retrieves current or historical observations~\cite{hou2026interlv} and robot states; \texttt{Reference} supplies optional demonstration context; and \texttt{Act} executes motion and gripper commands, returning updated observations and execution feedback.
\texttt{Status} checks the environment's task-completion predicate, while \texttt{Finish} terminates the interaction.

Specifically, Algorithm~\ref{alg:closed_loop} summarizes the workflow. Starting with the initial \texttt{Observe} and a\texttt{Reference} call, the model uses feedback from each action to continue execution, retrieve historical evidence or verify completion.
Successful environment verification leads to \texttt{Finish}; otherwise, execution resumes until completion or budget exhaustion.

\subsection{History State Retrieval}
\texttt{Observe} retrieves a stored observation record by its nonnegative ID $j$, providing temporal context for diagnosis and subsequent planning. We distinguish this record ID from the decision index $t$. For a record produced by an \texttt{Act} call at decision $t_j$, we store
\begin{equation}
    \text{Observe}(j) = \big(\bar{o}_j(\mathcal{M}),\,u(a)_{t_j},\,r_{t_j}\big),
\end{equation}
where $\bar{o}_j(\mathcal{M})$ denotes the post-action observation stored in record $j$, including the enabled optional inputs, and $r_{t_j}$ is the execution-feedback component of the tool result $z_{t_j}$. The initial record contains only the initial observation, with no preceding action or execution feedback.
Historical retrieval is read-only: it does not recapture images, advance the simulator, or consume environment control steps. The model can therefore recall past events without altering the physical state. Before issuing an action, it must inspect and reference the latest observation record.

\subsection{Embodied In-context Reference}
\texttt{Reference} supplies an optional, prevalidated demonstration package to support embodied in-context learning~\cite{cheng26incontext_robot_learning, liu2026roboicl}. The package may contain video keyframes, textual step descriptions, action trajectories, or a combination of these modalities. It is provided after the initial observation and before the first \texttt{Act} call, and incorporated into the interaction history $\mathcal{H}_t$.
Let $R$ denote the demonstration context. We make its role explicit in the tool-selection policy:
\begin{equation}
    u(a)_t \sim \pi_\theta\!\left(\cdot\mid\mathcal{H}_t,\ell,R\right),
\end{equation}
where $R$ is written separately to highlight the demonstration context already included in $\mathcal{H}_t$. It provides guidance on manipulation order and contact strategies, while the model determines motion parameters from current observations and execution feedback. This interface enables us to evaluate how demonstration context supports action generation and closed-loop adaptation.

\subsection{Action Execution Interface}
\label{sec:act}
\texttt{Act} converts a model decision into a bounded, synchronous control command. In the dual-arm interface, both arms are specified explicitly. For arm $k\in\{L,R\}$, the action arguments $a_t$ contain
\begin{equation}
    a_t^k = \left(\Delta p_t^k,\,\Delta q_t^k,\,g_t^k,\,M_t^k\right),
\end{equation}
where $\Delta p_t^k\in\mathbb{R}^3$ is a translation in meters in the controller's reference frame, $\Delta q_t^k$ is a unit quaternion specifying a rotation increment in that frame, $g_t^k\in[0,1]$ is the gripper target, and $M_t^k$ is the arm's independent limit on solver steps. Let $(p_{t,\mathrm{start}}^k,q_{t,\mathrm{start}}^k)$ denote the measured end-effector pose in the same frame at the start of the call. The target pose is
\begin{equation}
    p_{t,\mathrm{goal}}^k = p_{t,\mathrm{start}}^k+\Delta p_t^k, \qquad q_{t,\mathrm{goal}}^k = \Delta q_t^k\otimes q_{t,\mathrm{start}}^k,
\end{equation}
where $\otimes$ denotes quaternion composition, with the increment applied by left multiplication. The controller executes these targets through bounded motion increments, synchronizing the two arms while allowing each to have an independent execution limit. An arm that reaches its target maintains that target as the other continues. Each synchronized step is checked against kinematic and joint constraints and rejected for both arms if either command is invalid.

After execution, the tool returns updated observations and feedback $r_t$ as part of $z_t$. Motion completion and residual pose errors provide evidence for the model to assess the attempted action and revise subsequent commands. This feedback connects model-generated geometric decisions to their physical outcomes, enabling closed-loop correction. Task-dependent motion planning remains the model's responsibility; the controller's feasibility checks do not guarantee collision-free motion or successful grasping.

\section{Experiments}
We evaluate GPT-6 Astra on robot manipulation along decomposed complementary dimensions. Our systematic experiments cover four aspects: depth perception, embodied in-context learning, tactile understanding, and integrated manipulation performance.
For depth perception, we test whether the model performs metric reasoning about 3D space or only relies on 2D visual cues. 
For embodied in-context learning, we examine whether additional demonstrations transfer task intent or support fine motor skills. 
For tactile understanding, we determine whether the model interprets tactile signals and adjusts force-dependent actions. 
Finally, we evaluate integrated manipulation performance for comparison, which probes the capability ceiling in complex and diverse scenes.

\subsection{Datasets}
\noindent\textbf{LIBERO-Plus}~\cite{fei25libero-plus}
extends LIBERO~\cite{liu2023libero} with a large-scale robustness evaluation suite for vision-language-action models. It contains 10,030 perturbed tasks based on LIBERO assets spanning 7 dimensions: object layout, camera viewpoint, robot initial state, language instruction, lighting, background texture, and sensor noise. 

\noindent\textbf{RoboDojo}~\cite{chen2026robodojo}
is a unified simulation-and-real benchmark for generalist manipulation policies. It includes 42 simulation tasks and 18 real-world tasks. The simulation suite evaluates generalization, memory, precision, long-horizon execution, and open-vocabulary instruction following, while the real-world suite tests deployment under physical interaction and standardized hardware protocols.

\noindent\textbf{RoboTwin 2.0}~\cite{chen2025robotwin2}
is a bimanual manipulation benchmark built with generative digital twins. It provides 50 bimanual tasks with strong domain randomization and supports scalable expert-data generation. The benchmark emphasizes dual-arm coordination, object interaction, and robustness to scene and embodiment variation.

\noindent\textbf{UniVTAC}~\cite{chen2026univtac}
is a simulation and benchmarking platform for visuo-tactile manipulation. It supports multiple tactile sensor types and provides scalable generation of paired visual and tactile interaction data. Its benchmark contains 8 tasks covering pose reasoning, shape perception, and contact-rich insertion.

\noindent\textbf{RoboCasa GR1}~\cite{bjorck2025gr00t}
is a tabletop manipulation benchmark built on the RoboCasa simulation framework~\cite{robocasa2024}. It comprises 24 tasks performed by a Fourier GR-1 humanoid robot equipped with dexterous hands, covering object placement and interactions with household fixtures.

\subsection{Implementation Details}
GPT-6 Astra is the only planning component for evaluation. No other robot-specific models are adopted. All experiments are conducted through Codex with the reasoning effort set to high. Each \texttt{Act} call specifies a relative pose target that the controller tracks over multiple native control steps. The translation and rotation increments are limited to 0.01\,m and 0.05\,rad per step, respectively. The requested displacement norm and shortest rotation angle are bounded by 0.20\,m and $\pi/2$\,rad, respectively. By default, each \texttt{Act} call permits up to 10 native control steps.

\subsection{Evaluation Metrics}
To provide a comprehensive assessment of GPT-6 Astra, we use 2 task effectiveness metrics and 2 efficiency metrics. They capture not only whether the model completes each task, but also the interaction and reasoning effort it requires.

\begin{itemize}
    \item \textbf{Success Rate (SR).} SR is the percentage of trials that satisfy the benchmark-specific success condition. Partial progress is not counted as success.

    \item \textbf{Score.} Score is the official process-oriented metric for the benchmark RoboDojo~\cite{chen2026robodojo}, computed as the mean process reward multiplied by 100. It credits partial task progress and provides a finer-grained comparison among failed episodes.
    
    \item \textbf{Action Rounds.} Action rounds denote the number of model-to-controller action proposals issued in a trial. This metric measures interaction efficiency.
    
    \item \textbf{Total Tokens.} Total tokens denote the cumulative number of input and output tokens consumed during a trial. This metric measures the API usage of robotic manipulation.
    
    
\end{itemize}

\subsection{Capabilities Evaluation}
\label{sec:capabilities_evalution}
\begin{table}[ht]
\centering
\caption{\textbf{Depth perception ablation with representative tasks from benchmark LIBERO-Plus.} The Action Rounds and Total Tokens are averaged over all evaluated episodes.} 
\label{tab:depth_ablation_overall}
\resizebox{\textwidth}{!}{
\begin{tabular}{lcccc}
\toprule
Observation & Success Rate (\%) & Action Rounds & Total Tokens \\
\midrule
w/ RGB                                                  & 47.5 & 34.61 & 1,046,755  \\
\midrule
w/ RGB + Camera parameters                              & 60.0 & 26.32 & 986,066  \\
w/ RGB + Camera parameters + Explicit depth estimation  & 62.5 & 28.85 & 1,119,366 \\
w/ RGB + Depth map                                      & 67.5 & 24.23 & 1,101,271 \\
w/ RGB + Back-projection tool                           & 60.0 & 27.00 & 1,377,672  \\
\bottomrule
\end{tabular}
}
\end{table}

\begin{figure*}
    \centering
    \includegraphics[width=\textwidth, trim=0mm 0mm 0mm 0mm]{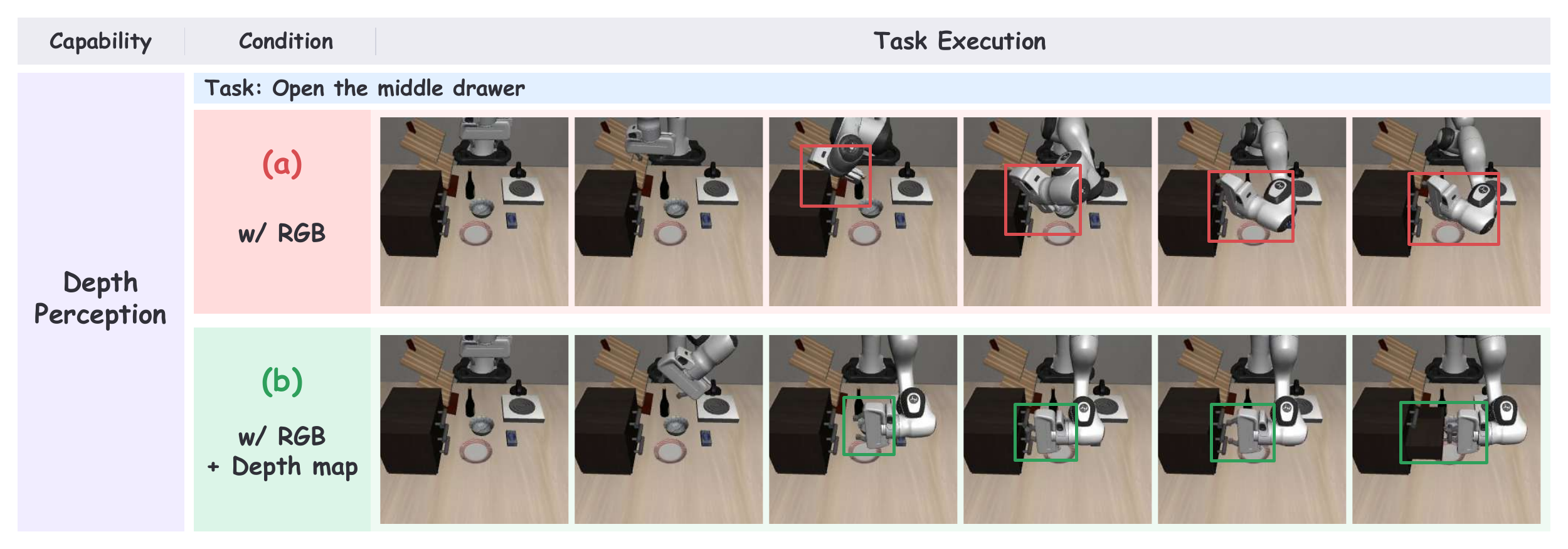}
    \caption{\textbf{The Visualization of depth perception ablation.} The selected task is \textit{open the middle drawer of the cabinet} from benchmark LIBERO-Plus. Green grounding boxes highlight successful actions, while red grounding boxes highlight failure-inducing actions.}
    \label{fig:depth_perception_visualization}
\end{figure*}

\subsubsection{Depth Perception}
The depth ablation study examines whether GPT-6 Astra can understand depth information and use it to reason about physical interactions from a 3D spatial perspective~\cite{chen2024spatialvlm,huang2023voxposer}. Specifically, we select four demanding tasks from LIBERO-Plus and conduct ten trials per task under each configuration. To investigate how different forms of depth-related information affect performance, we compare five configurations as summarized in Table~\ref{tab:depth_ablation_overall}:
\begin{itemize}
    \item \textbf{Baseline}: Task instruction and current RGB observations.
    \item \textbf{RGB + Camera parameters}: The RGB observations paired with camera intrinsic and extrinsic parameters.
    \item \textbf{RGB + Camera parameters + Explicit depth estimation}: The RGB observations paired with camera parameters and the model is required to explicitly estimate the depth of the centers of objects it interacts with before acting.
    \item \textbf{RGB + Depth map}: The RGB observations paired with a dense depth map rendered from ground-truth.
    \item \textbf{RGB + Back-projection tool}: An additional tool that returns the accurate 3D world coordinates of a queried 2D image point.
\end{itemize}

\begin{figure*}[t]
    \centering
    \includegraphics[width=\textwidth]{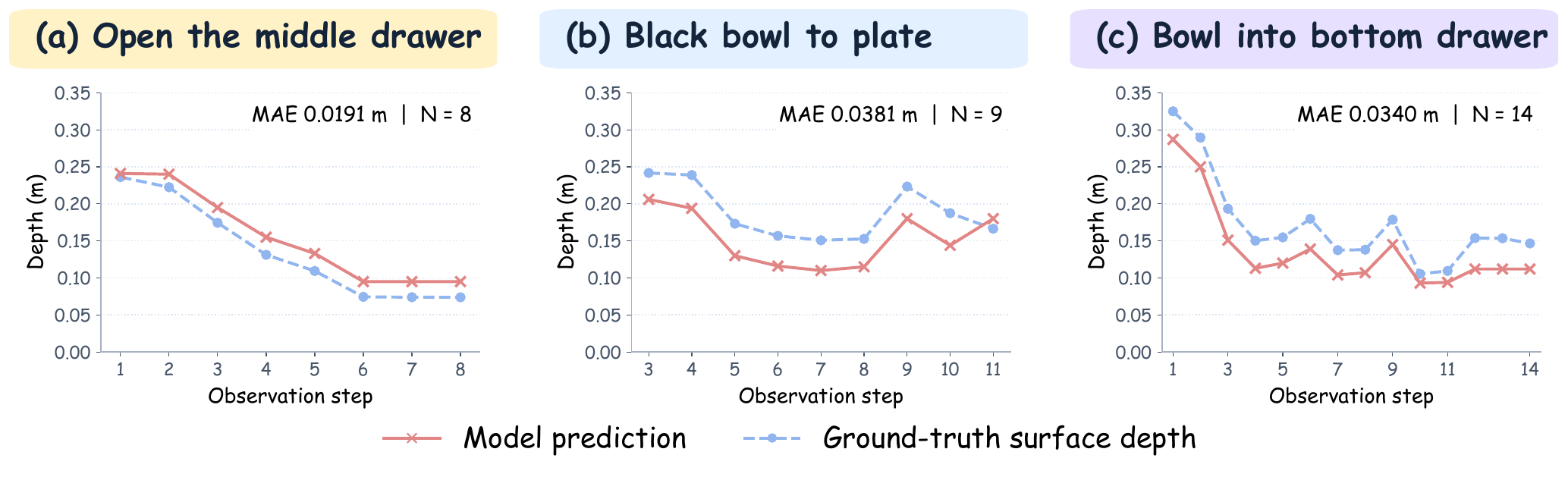}
    \caption{\textbf{Depth estimates compared with ground-truth depth.} Predictions and ground truth are shown across observation steps for three task examples. MAE denotes mean absolute error, and $N$ is the number of observations shown.}
    \label{fig:depth_estimation_accuracy}
\end{figure*}

\noindent\textbf{Camera Parameters Improve Manipulation.}
Adding camera parameters increases the success rate from 47.5\% with RGB alone to 60.0\% and reduces average action rounds from 34.61 to 26.32. This suggests that camera parameters help the model relate image positions to positions in the robot workspace, even without explicit depth values. To examine the model's depth estimation ability, we further require it to estimate the depth of target object centers before acting. This raises the success rate slightly to 62.5\%. Although explicit depth estimation slightly increases action rounds and token usage compared with camera parameters alone, the recorded estimates allow us to directly examine the model's depth estimation errors. 
In particular, figure~\ref{fig:depth_estimation_accuracy} compares these estimates with ground-truth surface depth in three task examples. The predicted depths generally follow the changes in ground truth across observation steps, with mean absolute errors of 1.91, 3.81 and 3.40 cm, respectively. These examples suggest that GPT-6 Astra can recover useful metric depth information from RGB observations and camera parameters, with average errors of a few centimeters.


\noindent\textbf{Dense Depth Improves Success and Efficiency.}
Providing dense ground-truth depth maps alongside RGB observations achieves the highest success rate (67.5\%) and the fewest average action rounds (24.23) among the tested configurations, with only a modest increase in token usage over RGB alone. 
Figure~\ref{fig:depth_perception_visualization} illustrates this benefit through the task \textit{open the middle drawer of the cabinet}, where the model must locate and approach the drawer handle. Without additional depth cues, the model commands an excessively large movement toward the handle. 
The arm collides with the cabinet corner, leaving the wrist misaligned. In contrast, access to scene depth helps the model select a suitable approach distance and reach the handle accurately for grasping. 
This example suggests that depth information supports more accurate approach movements and can reduce the need for corrective actions, consistent with the lower average action rounds observed across the evaluated tasks.

\noindent\textbf{Point Queries Underperform Dense Depth Maps.}
Compared with the depth-map setting, the back-projection tool yields a lower success rate of 60.0\% and higher average token usage of 1.38M per episode. This is 31.6\% higher than the RGB-only baseline and 25.1\% higher than the depth-map setting. 
These results suggest that accurate point coordinates alone are less helpful than dense depth maps in the tested settings. A depth map shows the depth of surrounding surfaces across the image, whereas each tool query returns only the position of a selected point. The model must still choose useful points and relate their coordinates to the objects and available space in the scene. This may require additional queries and reasoning, helping explain the higher token usage.

\noindent\textbf{Depth Alone Does Not Ensure Success.}
Overall, the depth ablations suggest that GPT-6 Astra can use camera parameters to improve manipulation, while explicitly estimating depth provides a modest additional gain. 
Camera parameters help the model relate visual observations to the robot workspace, and ground-truth depth maps further improve success and reduce action rounds. 
Dense depth maps also outperform the back-projection tool in the tested settings, suggesting that how depth information is provided matters. 
However, some tasks remain difficult even with ground-truth depth. This indicates that depth is not the only factor limiting task success.

\begin{table}[ht]
\centering
\caption{\textbf{Embodied in-context-learning ablation with representative tasks from dataset LIBERO-Plus.} The Action Rounds and Total Tokens are averaged over all evaluated episodes. All the contexts are from expert trajectories for the same task under different scene perturbations.}
\label{tab:icl_ablation_overall}
\resizebox{\textwidth}{!}{
\begin{tabular}{lccc}
\toprule
Context Formulation & Success Rate (\%) & Action Rounds & Total Tokens \\
\midrule
w/o context                                          & 64.0 & 23.00 & 1,187,814 \\
\midrule
w/ First and last frames                              & 46.0 & 21.18 & 1,448,025 \\
w/ Video sequence frames                                & 64.0 & 20.72 & 1,442,084 \\
w/ Textual step description                           & 78.0 & 21.30 & 1,052,921 \\
w/ Action trajectory + first and last frames              & 66.0 & 20.64 & 1,218,883 \\
w/ Action trajectory + video sequence frames            & 82.0 & 19.92 & 1,222,021 \\
w/ Action trajectory + textual step description       & 82.0 & 21.00 & 1,191,807 \\
w/ Action trajectory + video sequence frames + textual step description     & 86.0 & 20.58 & 1,372,547 \\
\bottomrule
\end{tabular}%
}
\end{table}

\begin{figure*}[ht]
    \centering
    \includegraphics[width=\textwidth, trim=0mm 10mm 0mm 10mm]{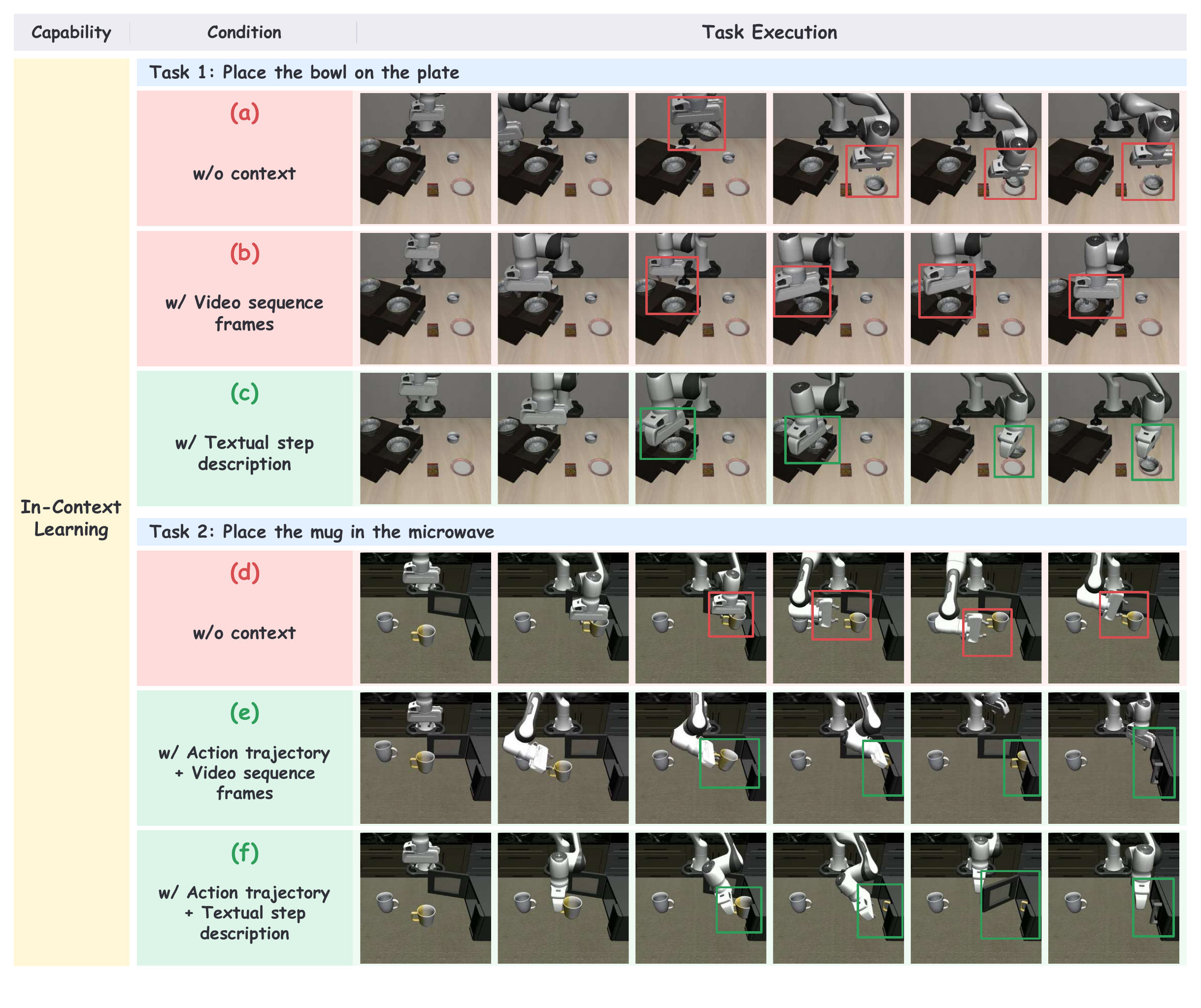}
    \caption{\textbf{The Visualization of in-context learning ablation.} The shown tasks are \textit{pick the akita black bowl in the top layer of the wooden cabinet and place it on the plate} and \textit{put the yellow and white mug in the microwave and close it} from dataset LIBERO-Plus. Green grounding boxes highlight successful actions, while red grounding boxes highlight failure-inducing actions.}
    \label{fig:icl_learning_visualization}
    \vspace{-5bp}
\end{figure*}


\subsubsection{Embodied In-Context Learning}
\label{sec:icl_experiments}

We investigate whether GPT-6 Astra can acquire task-specific manipulation behaviors from contextual demonstrations without parameter updates~\cite{cheng26incontext_robot_learning, liu2026roboicl}. We select five challenging tasks from LIBERO-Plus, with ten trials per task under each condition. To examine how different forms of demonstration information affect performance, we evaluate eight context formulations, as summarized in Table~\ref{tab:icl_ablation_overall}:

\begin{itemize}
    \item \textbf{Baseline}: Task instruction and current observations without demonstrations.
    \item \textbf{First and last frames}: Initial and final frames of a demonstration.
    \item \textbf{Video sequence frames}: Sampled frames showing intermediate states of a demonstration.
    \item \textbf{Textual step description}: Step-by-step descriptions of the manipulation procedure.
    \item \textbf{Action trajectory + first and last frames}: Action references paired with initial and final frames.
    \item \textbf{Action trajectory + video sequence frames}: Action references paired with sampled demonstration frames.
    \item \textbf{Action trajectory + textual step description}: Action references combined with procedural descriptions.
    \item \textbf{Action trajectory + video sequence frames + textual step description}: A combination of visual, textual, and action-level demonstrations.
\end{itemize}
\noindent\textbf{Visual Demonstrations Alone Provide Limited Benefits.}
The results show that visual demonstrations do not necessarily improve manipulation performance. Compared with the 64.0\% success rate without context, providing only the first and last frames reduces success to 46.0\%, while video sequence frames achieve 64.0\%, matching the baseline. These findings suggest that visual demonstrations can convey task goals and intermediate states but may not provide sufficient guidance for inferring feasible manipulation motions. Sparse visual references can even introduce ambiguity when the demonstrated scene differs from the current environment. Thus, simply increasing the amount of visual context does not guarantee improved execution.

\noindent\textbf{Textual Guidance Improves Task Understanding and Planning.}
In contrast, textual step descriptions increase the success rate from 64.0\% to 78.0\%, while reducing token consumption from 1.19M to 1.05M. This suggests that explicit procedural guidance helps the model decompose tasks and identify relevant manipulation constraints. Figure~\ref{fig:icl_learning_visualization} illustrates this distinction through the bowl-placement task. Without context, GPT-6 Astra mistakenly selects a bowl on top of the cabinet rather than the target inside the drawer. Both video frames and textual descriptions help identify the correct target, but video frames alone fail to guide a feasible grasp because the attempted motion is obstructed by the cabinet. In contrast, textual guidance leads to a suitable approach and successful placement. This example highlights that understanding the task goal is not sufficient; explicit procedural information can bridge the gap between target identification and successful execution.

\noindent\textbf{Action Trajectories Enable Motion-Level Adaptation.}
Beyond semantic guidance, action trajectories provide more direct information about manipulation behaviors. Adding trajectories to video frames improves success from 64.0\% to 82.0\%, while combining trajectories with textual descriptions increases success from 78.0\% to 82.0\%. Even when paired with only the first and last frames, trajectories improve success from 46.0\% to 66.0\%. Combining all three modalities achieves the highest success rate of 86.0\%. These consistent improvements suggest that action references contribute motion-level information that cannot be fully conveyed by visual observations or textual instructions alone.

The microwave task in Figure~\ref{fig:icl_learning_visualization} provides further evidence. Without demonstrations, GPT-6 Astra repeatedly attempts a top-down grasp, but the microwave door obstructs its motion. With action trajectory references, the model adjusts its arm orientation and adopts a feasible manipulation sequence, successfully completing the task. Importantly, the demonstrated actions are not directly replayed but adapted to the current scene. This behavior suggests that embodied ICL can support the acquisition and adaptation of simple motor strategies rather than merely improving semantic task interpretation.

\noindent\textbf{Overall Findings and Efficiency Trade-off.}
Together, these experiments demonstrate that the effectiveness of embodied ICL depends critically on how demonstration information is represented. Textual descriptions primarily support task understanding and procedural planning, while action trajectories further guide physical execution. Although the full multimodal configuration achieves the highest success rate of 86.0\%, it consumes 1.37M tokens, compared with 1.19M tokens for trajectory-plus-text guidance at 82.0\% success. This trade-off suggests that richer context is not always the most efficient choice. More importantly, the results provide evidence that GPT-6 Astra can benefit from contextual demonstrations at both semantic and action levels without additional training.

\begin{table}[!t]
\centering
\caption{\textbf{Tactile-understanding ablation with representative tasks from UniVTAC benchmark.} The Action Rounds and Total Tokens are averaged over all evaluated episodes.}
\label{tab:tactile_ablation}
\begin{tabular}{lccc}
\toprule
Observation & Success Rate (\%) & Action Rounds & Total Tokens \\
\midrule
w/ RGB & 14.0 & 10.57 & 423.90k \\
\midrule
w/ Tactile (Tactile-only) & 16.0 & 12.38 & 569.02k \\
w/ RGB + Tactile (image) & 30.0 & 10.53 & 646.06k \\
w/ RGB + Tactile (text) & 50.0 & 9.44 & 451.19k \\
w/ RGB + Tactile (image + text) & 48.0 & 9.75 & 553.36k \\
\bottomrule
\end{tabular}%
\end{table}

\begin{figure*}[!t]
    \centering
    \includegraphics[width=\textwidth]{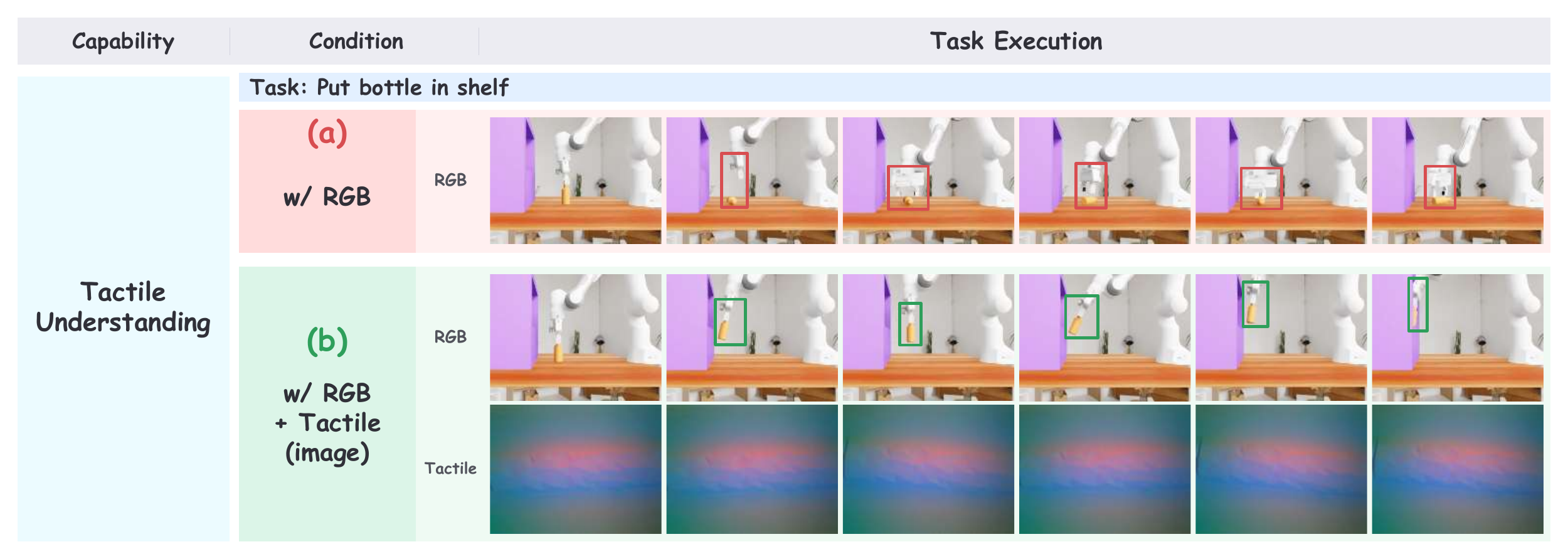}
    \caption{\textbf{The visualization of tactile understanding ablation.} The shown task is \textit{put bottle in shelf} from dataset UniVTAC. Green grounding boxes highlight successful actions, while red grounding boxes highlight failure-inducing actions.}
    \label{fig:tactile_understanding_rgb_tactile_image}
\end{figure*}

\begin{figure*}[!t]
    \centering
    \includegraphics[width=\textwidth, trim=0mm 0mm 0mm 0mm]{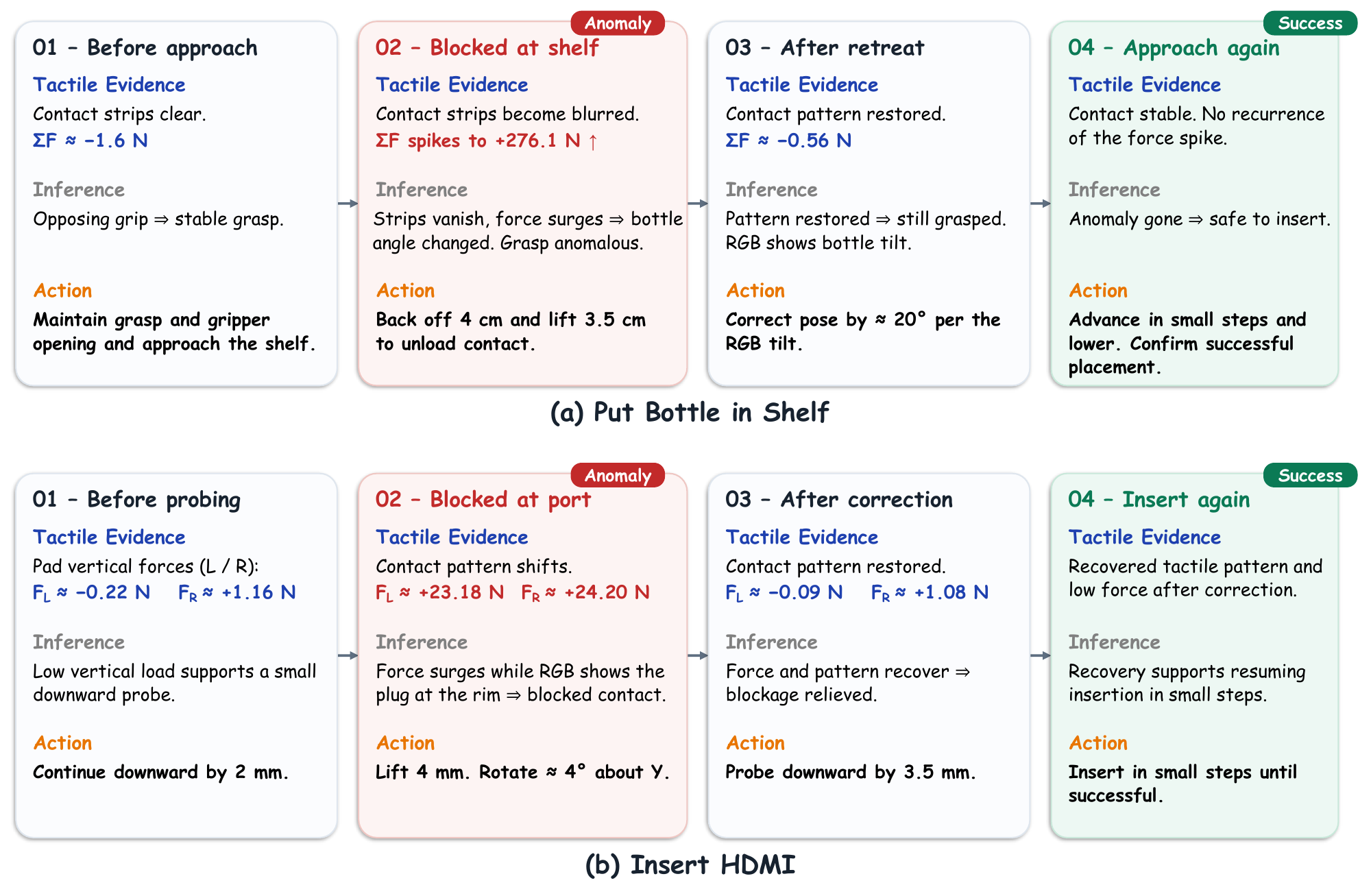}
    \caption{\textbf{The CoT chain of tactile understanding under the RGB + tactile (image + text) condition.} \(\Sigma F\) denotes the summed force in (a), while \(F_L\) and \(F_R\) denote the left and right pad vertical forces.}
    \label{fig:tactile_understanding_CoT}
\end{figure*}

\subsubsection{Tactile Understanding}
\label{sec:tactile_experiments}

We evaluate GPT-6 Astra on five representative UniVTAC~\cite{chen2026univtac} tasks with 20 trials per task to assess how tactile representations affect contact reasoning and force-sensitive manipulation. We compare five observation conditions, as summarized in Table~\ref{tab:tactile_ablation}:

\begin{itemize}
    \item \textbf{RGB}: Visual observations only, without tactile information.
    \item \textbf{Tactile-only}: Tactile signals only, without visual observations.
    \item \textbf{RGB + Tactile (image)}: RGB observations supplemented with raw tactile sensor images.
    \item \textbf{RGB + Tactile (text)}: RGB observations supplemented with structured textual descriptions of the measured force components along the x-, y-, and z-axes.
    \item \textbf{RGB + Tactile (image + text)}: RGB observations combined with both tactile images and textual force descriptions.
\end{itemize}

\noindent\textbf{Combining Vision and Touch Improves Manipulation Performance.} The results show that combining visual and tactile observations improves manipulation performance over either modality alone. RGB-only and tactile-only observations achieve success rates of 14.0\% and 16.0\%, respectively, while combining RGB with tactile images raises success to 30.0\%. Tactile-only also requires the most action rounds (12.38) and 34.2\% more tokens than RGB-only, suggesting that contact feedback without visual localization may lead to more exploratory interaction. Figure~\ref{fig:tactile_understanding_rgb_tactile_image} illustrates this distinction through the bottle-placement task. With RGB alone, the model repeatedly fails to grasp the bottle. In contrast, adding tactile images enables a successful grasp on the first attempt and subsequent placement on the shelf. This example suggests that contact feedback can complement visual observations in establishing a stable grasp~\cite{li2020tactile}.

\noindent\textbf{Textual Force Descriptions Improve Success and Efficiency.} Beyond the availability of tactile feedback, its representation also affects performance. Replacing tactile images with textual force descriptions increases success from 30.0\% to 50.0\%, while reducing average action rounds from 10.53 to 9.44 and token consumption from 646.06k to 451.19k. These results suggest that explicit force values provide more effective guidance for contact reasoning and corrective action selection in the tested settings. Unlike natural images, tactile images encode taxel responses rather than scene appearance, which may make them harder for the model to interpret. Textual descriptions expose the measured force components directly, potentially reducing the difficulty of interpreting the sensor response before selecting an action.



\noindent\textbf{Tactile Feedback Supports Contact Reasoning and Action Correction.} 
The examples in Figure~\ref{fig:tactile_understanding_CoT} provide further evidence of how tactile feedback supports action correction under the \emph{RGB + tactile (image + text)} condition. In \emph{Put Bottle in Shelf}, a sharp increase in resultant force prompts retreat and unloading, while subsequent force recovery supports renewed approach after pose correction. Similarly, in \emph{Insert HDMI}, increased vertical forces indicate blocked contact, and their reduction after correction supports resuming insertion. These behaviors suggest that the model can relate changes in force measurements to contact conditions and adjust its actions accordingly.

However, providing both tactile representations does not necessarily improve execution. Combining tactile images and text achieves 48.0\% success, compared with 30.0\% for tactile images alone and 50.0\% for tactile text alone, while consuming more tokens than the latter. Together with the illustrated reliance on force measurements, these results suggest that numerical and symbolic reasoning plays an important role in the model's use of tactile feedback. The evaluated combination provides no clear additional control benefit over textual force descriptions alone.

\noindent\textbf{Overall Findings and Remaining Limitations.}
Overall, these experiments suggest that GPT-6 Astra can use tactile feedback to support both grasping and contact-driven action correction. As with depth, the effectiveness of this feedback depends on how embodied information is represented, with textual force descriptions providing the strongest results here. Nevertheless, substantial failures remain even in the best-performing condition, indicating that the ability to reason about contact does not yet ensure reliable force-sensitive manipulation.

\begin{figure*}[ht]
    \centering
    \includegraphics[width=\textwidth]{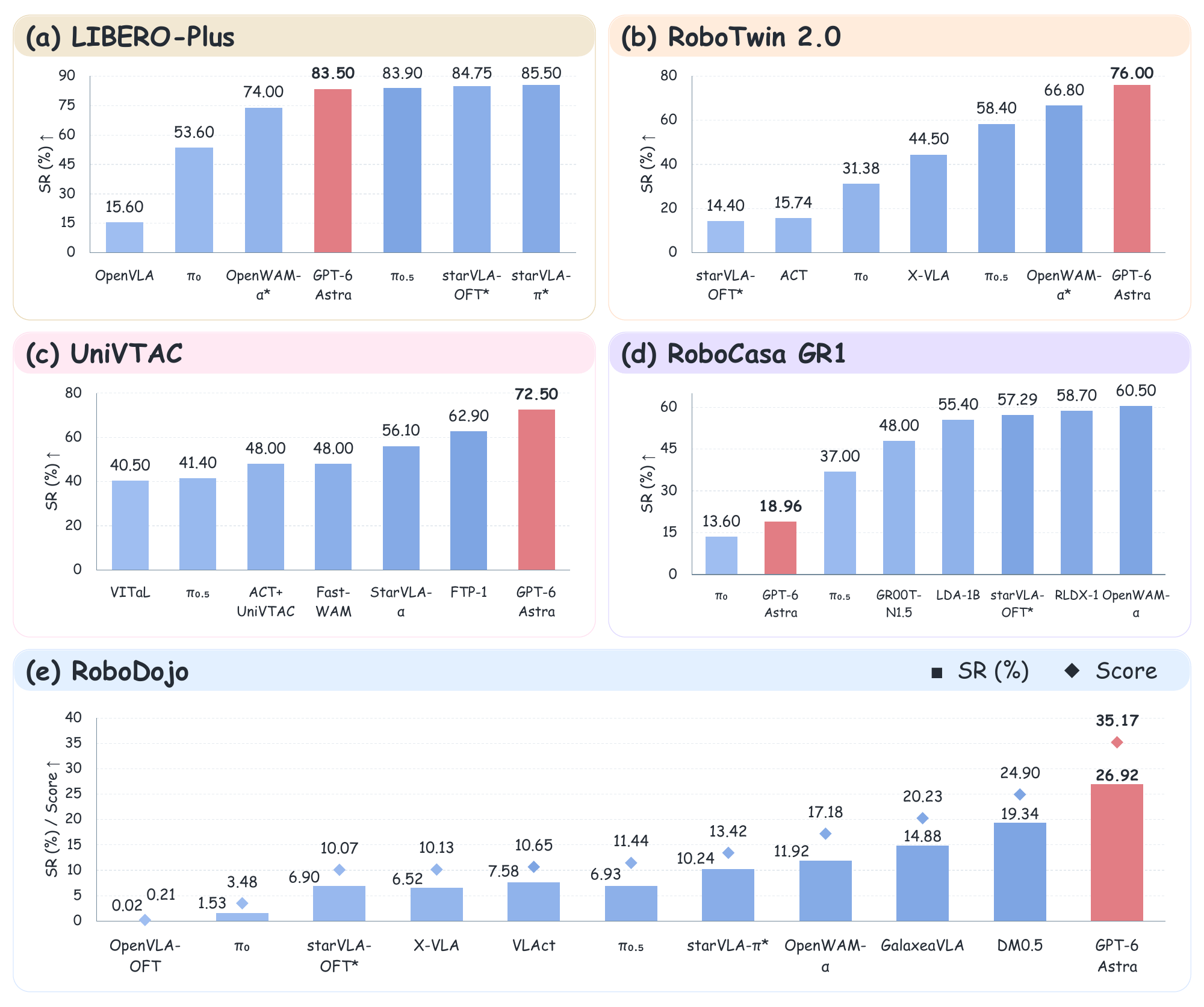}
    \caption{\textbf{The comprehensive comparison of GPT-6 Astra with existing solutions}~\cite{kim2025openvla,black2025pi0,wang2026openwam,physicalintelligence2025pi05,community2026starvla,zhao2023act,zheng2026xvla,george2025vital,chen2026univtac,yuan2026fastwam,ye2026starvla,yuan2026ftp1,nvidia2025gr00tn15,lyu2026lda1b,kim2026rldx1,kim2025openvlaoft,yang2026vlact,jiang2025galaxea,dexmal2026dm05, fei25libero-plus}\textbf{.} Bars indicate success rate (SR), and diamonds additionally report the process-oriented score on RoboDojo. Higher values indicate better performance. * represents the reevaluated methods with our protocol. GPT-6 Astra is highlighted with red.}
    \label{fig:comparison}
\end{figure*}

\subsection{Main Results}
\label{sec:main_results}


\noindent\textbf{Evaluation setting.}
Following the capability-level analyses, we evaluate GPT-6 Astra on complete manipulation tasks across five benchmarks to examine how its embodied capabilities jointly support closed-loop execution. 
For consistency with subsequent real-world experiments in section~\ref{sec:real_world}, we provide only RGB observations and camera parameters for depth perception, requiring the model to explicitly estimate target object depths before acting. 
For embodied in-context learning, we provide demonstrations combining sampled video frames, textual step descriptions, and action trajectories. 
Tactile feedback and ground-truth depth map are additionally provided only on UniVTAC. These capabilities are integrated within our proposed agent execution framework as shown in Figure~\ref{fig:framework}.

We evaluate full tasks in each of the five benchmarks, covering 164 tasks in total: 40 base tasks from LIBERO-Plus, 50 from RoboTwin 2.0, 42 from RoboDojo, 24 from RoboCasa GR1, and 8 from UniVTAC. Given the computational cost of closed-loop evaluation, we use a reduced set of trials rather than the full official evaluation protocols. Each task is evaluated with 20 trials on RoboCasa GR1 and UniVTAC, and 10 trials on the other three benchmarks. 
To assess the consistency of this protocol, we reevaluate a subset of selected existing methods. For these methods, the resulting success rates remain close to their officially reported values, which suggests that performance estimates are reasonably stable despite the reduced evaluation budget. 
Figure~\ref{fig:comparison} compares GPT-6 Astra with existing methods, using our reevaluated results for a subset of baselines and officially reported results for the others. These comparisons provide an empirical reference for GPT-6 Astra's integrated manipulation capabilities across the evaluated benchmarks.

\noindent\textbf{Competitive integrated performance.}
On LIBERO-Plus shown in Figure~\ref{fig:comparison} (a), GPT-6 Astra achieves an 83.50\% success rate, showing competitive performance relative to the baselines included in our comparison. This is close to the officially reported 83.90\% of $\pi_{0.5}$~\cite{physicalintelligence2025pi05}. Our reevaluated starVLA-OFT and starVLA-$\pi$~\cite{community2026starvla} achieve slightly higher success rates of 84.75\% and 85.50\%, respectively. LIBERO-Plus introduces variations in scene layout, camera viewpoint, language, and other observation conditions. This result suggests that GPT-6 Astra can adapt task interpretation and spatial grounding to varied inputs within a closed loop.

On RoboTwin 2.0 shown in Figure~\ref{fig:comparison} (b), GPT-6 Astra reaches 76.00\%, the highest success rate among the methods shown. It exceeds our reevaluated OpenWAM-$\alpha$~\cite{wang2026openwam}, which ranks second in this comparison, by 9.20 percentage points. The benchmark emphasizes bimanual manipulation under scene variation. This result suggests that GPT-6 Astra can translate task understanding into coordinated actions across two arms, extending its utility beyond single-arm manipulation.

On UniVTAC shown in Figure~\ref{fig:comparison} (c), GPT-6 Astra achieves 72.50\%, markedly outperforming the reported 62.90\% of FTP-1~\cite{yuan2026ftp1} and 56.10\% of starVLA. The absolute gains are 9.60 and 16.40 percentage points, respectively. This benchmark includes pose reasoning, shape perception, and contact-rich insertion. Together with our tactile analysis, this result suggests that GPT-6 Astra can use visual and contact evidence to guide interaction when visual localization alone is insufficient.

On RoboDojo shown in Figure~\ref{fig:comparison} (e), GPT-6 Astra obtains a 26.92\% success rate and a score of 35.17. Both are the highest values in the comparison, exceeding the reported 19.34\% and 24.90 of DM0.5~\cite{dexmal2026dm05}. RoboDojo spans generalization, memory, precision, long-horizon execution, and open-vocabulary instruction following. Its favorable aggregate performance suggests that GPT-6 Astra's capabilities remain useful across a diverse set of manipulation demands. The score further indicates meaningful intermediate progress, even when tasks are not completed. Despite its favorable relative performance, both absolute metrics remain low. GPT-6 Astra therefore has substantial room for improvement on this highly challenging benchmark.

\noindent\textbf{Limits of cross-benchmark transfer.}
The excellent results above do not extend uniformly to RoboCasa GR1 shown in Figure~\ref{fig:comparison} (d). GPT-6 Astra achieves only 18.96\% success on this benchmark. It exceeds the reported 13.60\% of $\pi_0$~\cite{black2025pi0}, but falls well below the reported 37.00\% of $\pi_{0.5}$. The gap also persists against our reevaluated starVLA-OFT and reported OpenWAM-$\alpha$, which reach 57.29\% and 60.50\%, respectively.

The other evaluated settings use grippers, whereas RoboCasa GR1 uses dexterous hands. These hands have substantially more degrees of freedom, making manipulation a higher-dimensional reasoning problem. The model must account for hand pose, finger configuration, and multiple contact relationships to form and maintain stable grasps. This places greater demands on GPT-6 Astra's physical and spatial understanding, as well as its ability to translate that understanding into coordinated control. Policies trained on demonstrations from this embodiment can acquire corresponding motor priors, while GPT-6 Astra must rely on general reasoning and online feedback. We hypothesize that these additional demands contribute to its weaker performance, although aggregate results alone cannot establish the cause.

Overall, GPT-6 Astra demonstrates strong integrated manipulation capabilities across several benchmarks, with particularly notable zero-shot transfer. 
The compared policies learn manipulation from robot data within their respective training distributions before evaluation. In contrast, GPT-6 Astra operates without benchmark-specific policy training or parameter updates. Its competitive performance under zero-shot setting highlights the potential of general-purpose models for embodied control. 
This strength coexists with clear limitations. Success rate remains low on RoboDojo, and performance falls substantially behind trained policies on RoboCasa GR1. 
Although general reasoning provides a strong basis for manipulation, it doesn't ensure reliable embodiment-specific motor control. These benchmark-level trends provide a basis for further analysis of the capabilities and failure modes underlying GPT-6 Astra's manipulation performance.

\begin{table*}[!t]
\centering
\small
\setlength{\tabcolsep}{4pt}
\renewcommand{\arraystretch}{1.12}
\caption{The statistics of six capability dimensions and assigned tasks across the five benchmarks. Long task names are abbreviated. There are a total of 164 tasks and every task can count toward multiple dimensions. Counts include all tasks assigned to each dimension through a primary or secondary requirement, beyond the displayed examples. }
\label{tab:systematic_evidence}
\begin{tabular}{@{}p{0.15\textwidth}p{0.21\textwidth}>{\centering\arraybackslash}p{0.07\textwidth}p{0.52\textwidth}@{}}
\toprule
Dimension & Operational requirement & Number & Representative tasks \\
\midrule
\textbf{Generalization} & Adapt to changes in objects, scene layouts, or observation conditions. & 120 & \raggedright
\textbf{LIBERO-Plus:} Soup to Basket; Milk to Basket, etc.\newline
\textbf{RoboTwin 2.0:} Pick Diverse Bottles; Pick Dual Bottles, etc.\newline
\textbf{RoboDojo:} Stack Blocks; Fold Clothes, etc.\newline
\textbf{RoboCasa GR1:} Novel Object: Plate to Box; Tray to Pot, etc.
\tabularnewline[3pt]
\midrule
\textbf{Memory} & Retain object identities, initial states, sequences or counts needed by later actions. & 6 & \raggedright
\textbf{RoboDojo:} Match and Pick from Conveyor; Cover Blocks, etc.
\tabularnewline[3pt]
\midrule
\textbf{Task Understanding} & Ground linguistic references, spatial relations, symbolic goals, and task rules. & 57 & \raggedright
\textbf{LIBERO-Plus:} Bowl beside Ramekin to Plate, etc.\newline
\textbf{RoboTwin 2.0:} Blocks Ranking RGB; Place A Left of B, etc.\newline
\textbf{RoboDojo:} General Pickup; Solve Equation, etc.\newline
\textbf{UniVTAC:} Grasp Classify.
\tabularnewline[3pt]
\midrule
\textbf{Long-horizon} & Complete a task through an extended sequence of interdependent subgoals. & 60 & \raggedright
\textbf{LIBERO-Plus:} Bowl to Bottom Drawer and Close, etc.\newline
\textbf{RoboTwin 2.0:} Handover Block; Stack Three Bowls, etc.\newline
\textbf{RoboDojo:} Organize Table; Fill Egg Holder, etc.\newline
\textbf{RoboCasa GR1:} Potato to Microwave and Close, etc.
\tabularnewline[3pt]
\midrule
\textbf{High-order operation} & Coordinate orientation changes, rotation, pouring, or motion around articulated joints. & 31 & \raggedright
\textbf{RoboTwin 2.0:} Rotate QR Code; Hanging Mug, etc.\newline
\textbf{RoboDojo:} Fasten Screws; Insert Key and Turn, etc.\newline
\textbf{UniVTAC:} Insert Hole (oblique hole), etc.\newline
\textbf{RoboCasa GR1:} Bottle to Cabinet and Close, etc.
\tabularnewline[3pt]
\midrule
\textbf{Precision} & Meet tight alignment, contact, insertion, support, or release tolerances. & 74 & \raggedright
\textbf{LIBERO-Plus:} Wine Bottle to Rack; Book to Caddy, etc.\newline
\textbf{RoboTwin 2.0:} Click Bell; Place Phone on Stand, etc.\newline
\textbf{RoboDojo:} Insert Tubes; Plug in Charger, etc.\newline
\textbf{UniVTAC:} Insert Tube; Insert HDMI, etc.
\tabularnewline[3pt]
\bottomrule
\end{tabular}
\end{table*}

\section{Systematic Analysis}
Beyond aggregate benchmark performance, we systematically characterize GPT-6 Astra's embodied capabilities across six task-requirement dimensions: \emph{Generalization}, \emph{Memory}, \emph{Task Understanding}, \emph{Long-horizon Execution}, \emph{High-order Operation}, and \emph{Precision}. Each task is assigned a primary label and, where applicable, secondary labels based on its requirements rather than its execution outcome. Table~\ref{tab:systematic_evidence} defines these dimensions and provides representative tasks. By combining dimension-wise performance with execution traces, we identify the model's relative strengths and investigate the failure modes underlying its limitations.

\subsection{Capability Profile}
\label{sec:overview_systematic_analysis}

Table~\ref{tab:dimension_success} summarizes GPT-6 Astra's performance across the six capability dimensions. Each task contributes to every dimension assigned to it and is counted once within each dimension.
The results reveal a clear disparity between semantic competence and physical execution. Tasks involving task understanding, generalization, and memory achieve aggregate success rates of 73.45\%, 59.20\%, and 56.67\%, respectively. In contrast, tasks requiring long-horizon execution, high-order operation, and precision achieve substantially lower rates of 33.03\%, 34.62\%, and 36.19\%. 
These results suggest that GPT-6 Astra is more effective at interpreting manipulation goals and adapting to task variations than reliably executing complex physical interactions. While these comparisons reflect differences in task requirements and difficulty rather than isolated model abilities, they reveal a recurring gap between understanding what to do and physically accomplishing it. To further investigate this gap, we examine representative execution sequences to identify the behaviors underlying successful manipulation and the failure modes that constrain its performance.

\begin{table*}[t]
\centering
\small
\setlength{\tabcolsep}{4pt}
\renewcommand{\arraystretch}{1.15}
\caption{\textbf{Success rates across all 164 tasks by capability dimension, using primary and secondary assignments.} Cells show success rate and successful/evaluated episodes. ''--'' indicates that the benchmark has no task assigned to that dimension.}
\label{tab:dimension_success}
\begin{tabular}{@{}p{0.15\textwidth}*{6}{p{0.125\textwidth}}@{}}
\toprule
Dataset & Generalization & Memory & Task\newline Understanding & Long-horizon & High-order\newline operation & Precision \\
\midrule
LIBERO-Plus & 83.50\%\newline (334/400) & -- & 87.43\%\newline (306/350) & 67.00\%\newline (67/100) & 60.00\%\newline (12/20) & 58.33\%\newline (35/60) \\
RoboTwin 2.0 & 76.00\%\newline (380/500) & -- & 91.67\%\newline (55/60) & 62.86\%\newline (88/140) & 58.89\%\newline (53/90) & 77.50\%\newline (124/160) \\
RoboDojo & 19.17\%\newline (23/120) & 56.67\%\newline (34/60) & 30.67\%\newline (46/150) & 17.33\%\newline (52/300) & 9.17\%\newline (11/120) & 15.71\%\newline (33/210) \\
RoboCasa GR1 & 22.22\%\newline (80/360) & -- & -- & 9.17\%\newline (11/120) & 7.50\%\newline (6/80) & 18.96\%\newline (91/480) \\
UniVTAC & -- & -- & 95.00\%\newline (19/20) & -- & 66.25\%\newline (53/80) & 69.29\%\newline (97/140) \\
\midrule
Overall & 59.20\%\newline (817/1380) & 56.67\%\newline (34/60) & 73.45\%\newline (426/580) & 33.03\%\newline (218/660) & 34.62\%\newline (135/390) & 36.19\%\newline (380/1050) \\
\bottomrule
\end{tabular}
\vspace{5bp}
\end{table*}

\begin{figure*}[!t]
    \vspace{5bp}
    \centering
    \includegraphics[width=\textwidth, trim=0mm 0mm 0mm 0mm]{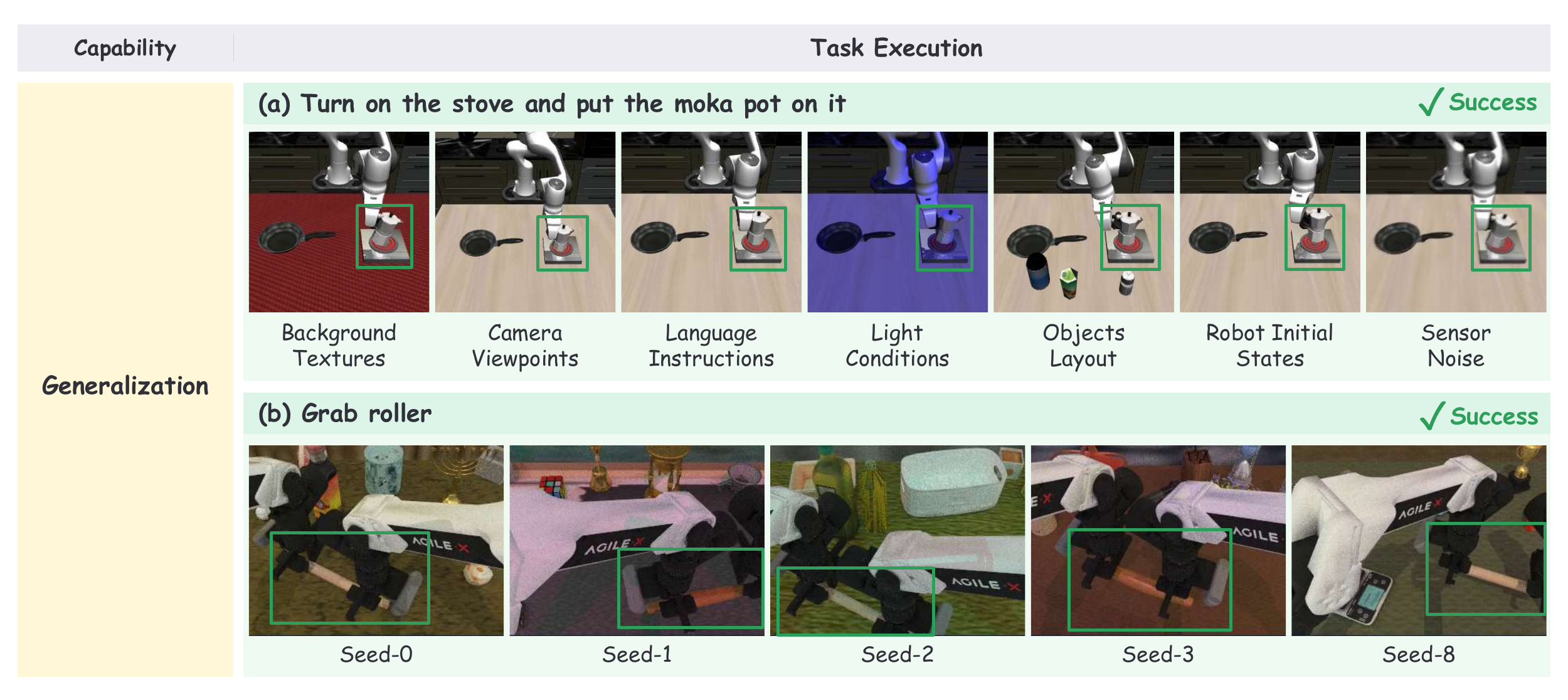}
    \caption{\textbf{The visualization of representative task executions demonstrating \emph{generalization}. }Green grounding boxes highlight successful actions, while red grounding boxes highlight failure-inducing actions.}
    \label{fig:generalization}
\end{figure*}

\begin{figure*}[!t]
    \vspace{5bp}
    \centering
    \includegraphics[width=\textwidth]{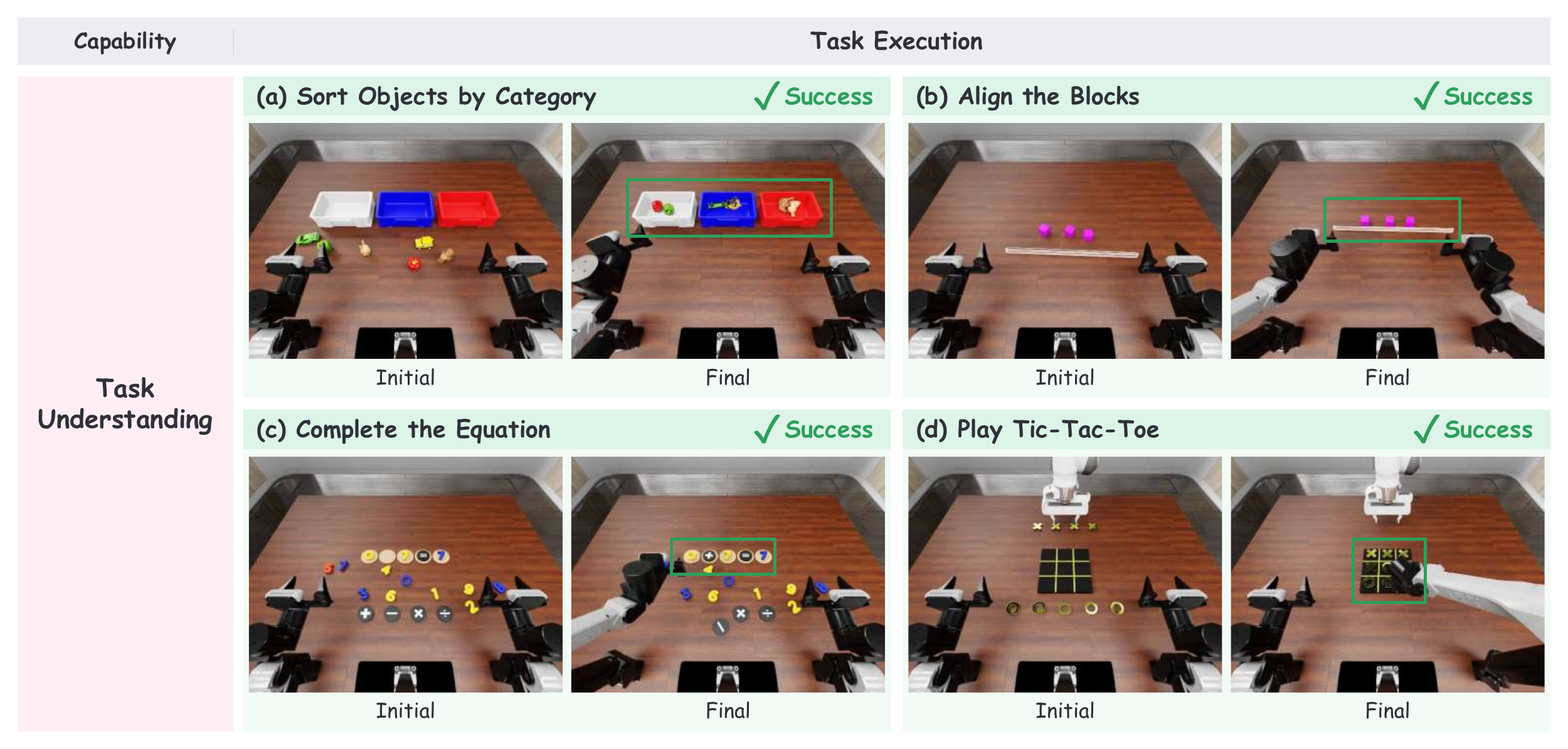}
    \caption{\textbf{The visualization of representative task executions demonstrating \emph{task understanding}.} Green grounding boxes highlight successful actions, while red grounding boxes highlight failure-inducing actions.}
    \label{fig:task_understanding}
\end{figure*}

\subsection{Strengths}
\label{sec:systematic_strengths}

\subsubsection{Generalization}

GPT-6 Astra demonstrates promising generalization across environmental and instructional variations without task-specific policy training. Figure~\ref{fig:generalization} illustrates this capability through representative tasks from LIBERO-Plus and RoboTwin 2.0. In Figure~\ref{fig:generalization} (a), the model successfully completes the stove-and-moka-pot task across seven types of perturbations, including changes in background textures, camera viewpoints, language instructions, lighting conditions, object layouts, robot initial states, and sensor noise. This demonstrates its ability to maintain consistent manipulation behavior despite variations in visual appearance and spatial configurations.
This robustness further extends to RoboTwin 2.0, where multiple scene factors vary simultaneously. As shown in Figure~\ref{fig:generalization} (b), GPT-6 Astra successfully grasps the roller across five different random seeds, adapting its actions to changing object configurations and surrounding environments.

Such robustness may stem from the broad visual and semantic knowledge acquired during general-purpose pretraining, enabling the model to interpret unfamiliar scene configurations and adapt its actions accordingly. Compared with specialized VLA and WAM policies, whose generalization remains closely tied to embodied training distributions, this suggests a promising advantage of leveraging general-purpose models for flexible robotic manipulation without task-specific policy training.

\subsubsection{Task Understanding}
GPT-6 Astra demonstrates strong task understanding, particularly when manipulation requires interpreting abstract instructions, implicit goals, and symbolic rules. Figure~\ref{fig:task_understanding} illustrates this capability through representative tasks from RoboDojo. 
In Figure~\ref{fig:task_understanding} (a), the model sorts objects according to their semantic categories, while in (b), it translates spatial relationships into concrete object arrangements. More challenging examples involve symbolic reasoning: completing an equation in (c) requires applying mathematical rules to select the correct symbol, while playing tic-tac-toe in (d) requires interpreting the board state and determining an appropriate move.

These examples demonstrate that GPT-6 Astra can infer task-relevant decisions beyond explicitly specified manipulation instructions, translating high-level goals into concrete object selections and spatial targets. This capability likely benefits from the broad semantic knowledge and reasoning abilities acquired through general-purpose pretraining, allowing the model to handle diverse task requirements without task-specific training. Such flexibility represents a key potential advantage of general-purpose models for open-ended robotic manipulation. However, correctly interpreting the intended goal does not necessarily guarantee successful physical execution, as further examined in our analysis of manipulation limitations.

\begin{figure*}[!t]
    \vspace{5bp}
    \includegraphics[width=\textwidth]{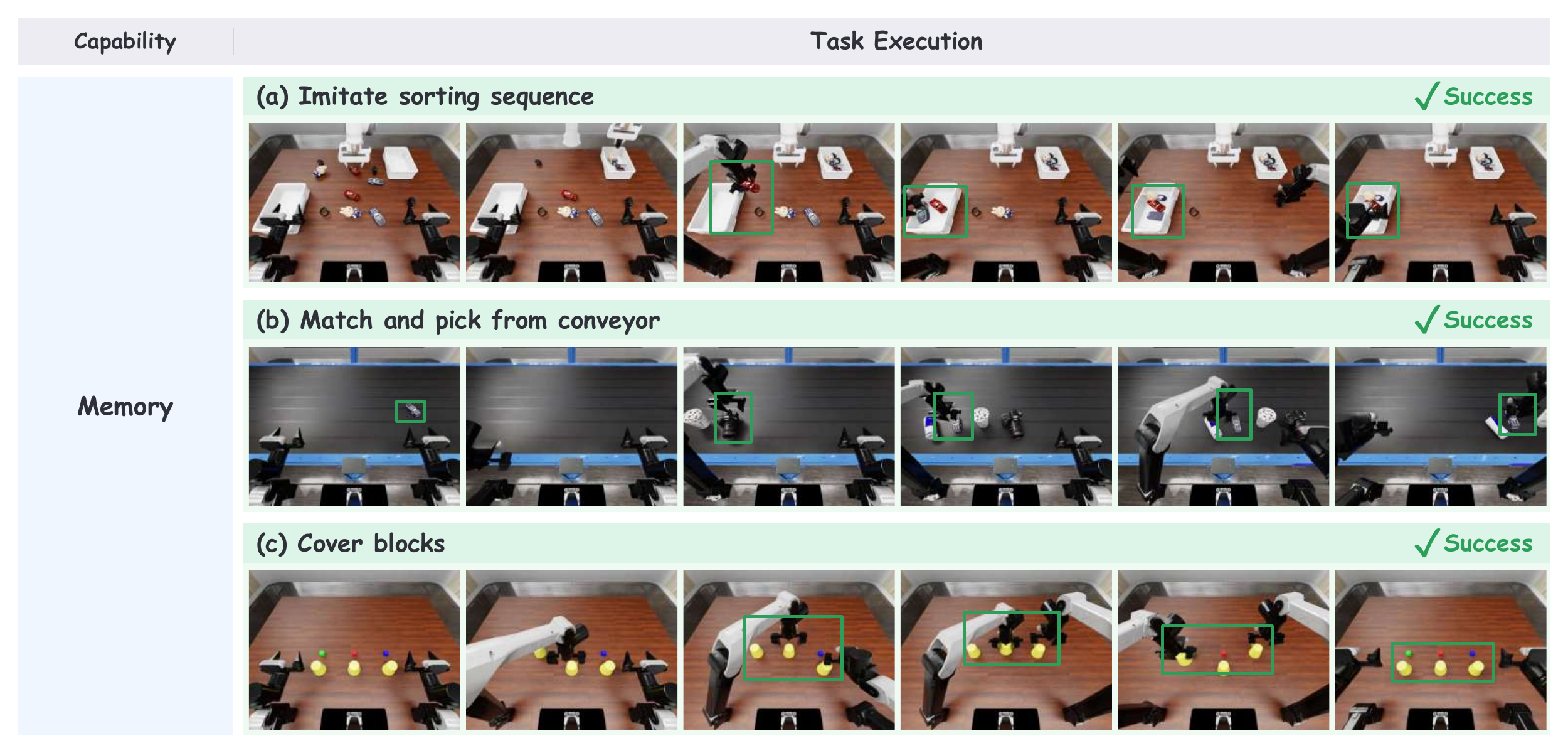}
    \caption{\textbf{The visualization of representative task executions demonstrating \emph{memory}.} Green grounding boxes highlight successful actions, while red grounding boxes highlight failure-inducing actions.}
    \label{fig:memory}
\end{figure*}

\subsubsection{Memory}
GPT-6 Astra demonstrates strong memory capabilities in manipulation, leveraging historical observations to guide actions when task-relevant information is no longer directly visible. Figure~\ref{fig:memory} illustrates this ability through three representative RoboDojo tasks involving sequence recall, target recognition, and spatial associations. In Figure~\ref{fig:memory} (a), the model reproduces a previously demonstrated sorting sequence, selecting and transferring objects to the tray in the demonstrated order. The conveyor task in Figure~\ref{fig:memory} (b) further examines memory under temporary occlusion. After observing a target object that subsequently disappears from view, the model successfully recognizes and grasps the same object among multiple candidates. Similarly, in the block-covering task shown in Figure~\ref{fig:memory} (c), the robot progressively covers colored blocks with cups, hiding their identities from the current observation. It later returns to these locations and removes the cups, demonstrating its ability to retain associations between object identities and spatial locations despite occlusion.

These observations highlight GPT-6 Astra's ability to reason over historical information rather than relying solely on the current visual state. This capability may benefit from the extensive context windows of general-purpose models, which can accommodate long interaction histories. Moreover, agentic harness complement this capability by preserving and organizing historical observations for subsequent retrieval. 
Together, long-context reasoning and structured history retrieval provide a promising foundation for manipulation tasks requiring information acquired earlier in execution. 
Nevertheless, effective memory does not necessarily guarantee reliable long-horizon manipulation, which additionally requires anticipating subgoal dependencies and coordinating actions over time.

\begin{figure*}[!t]
    \includegraphics[width=\textwidth, trim=0mm 0mm 0mm 0mm]{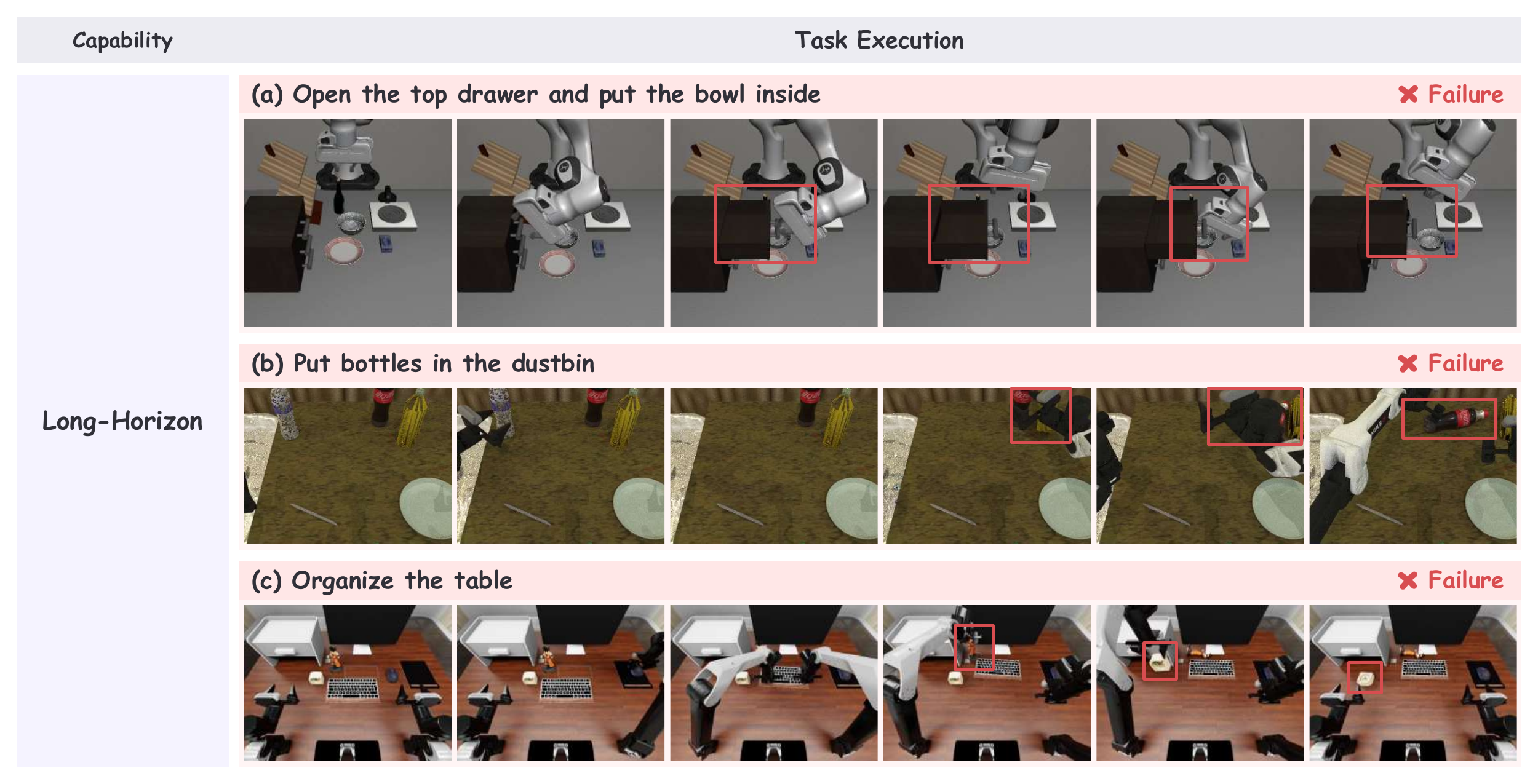}
    \caption{\textbf{The visualization of representative task executions illustrating limitations in \emph{long horizon}.} Green grounding boxes highlight successful actions, while red grounding boxes highlight failure-inducing actions.}
    \label{fig:long_horizon}
\end{figure*}

\begin{figure*}[!t]
    \includegraphics[width=\textwidth, trim=0mm 0mm 0mm 0mm]{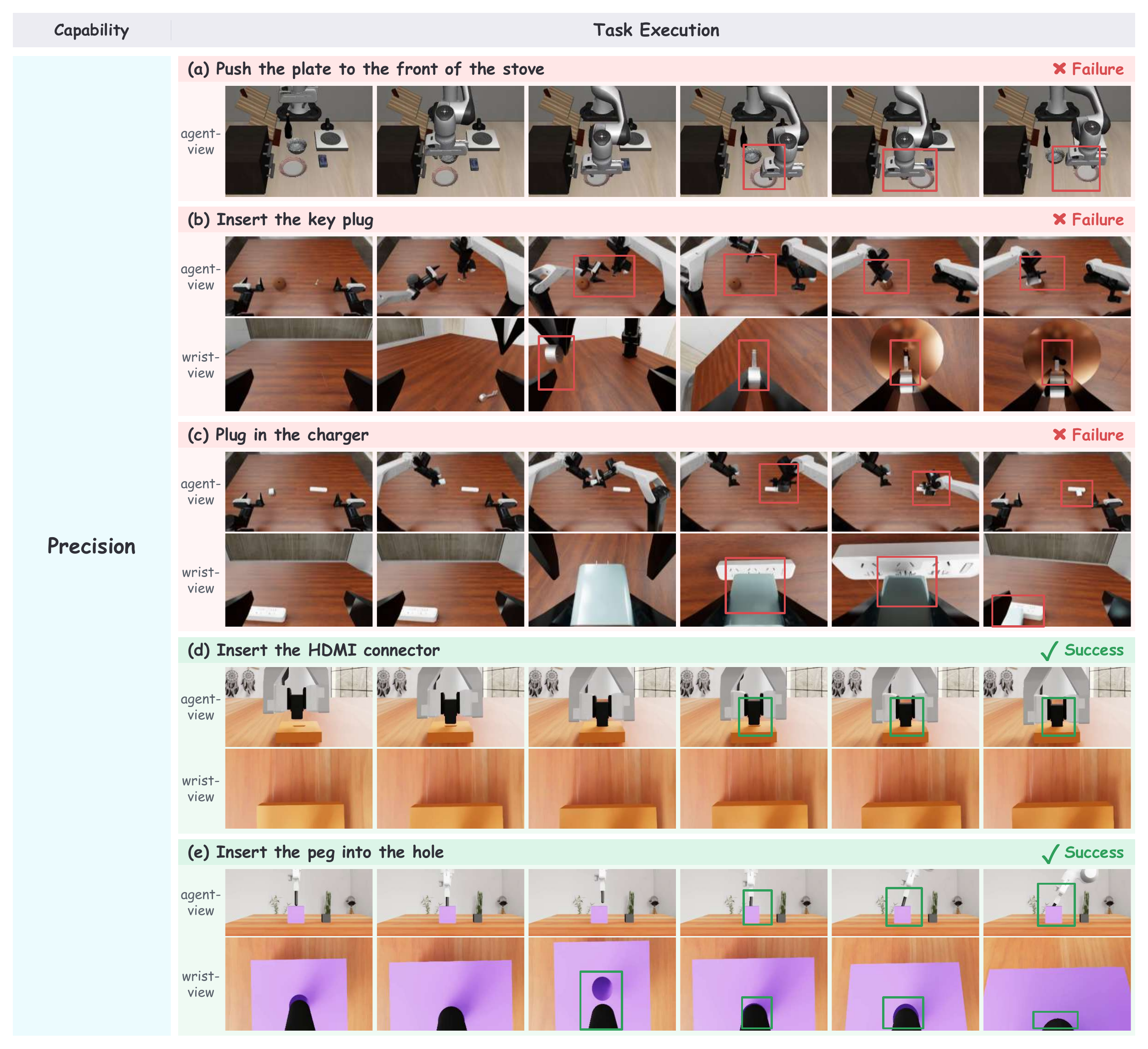}
    \caption{\textbf{The visualization of representative task executions illustrating limitations in \emph{precision}.} Green grounding boxes highlight successful actions, while red grounding boxes highlight failure-inducing actions.}
    \label{fig:precision}
\end{figure*}

\subsection{Limitations}
\label{sec:systematic_limitations}
\subsubsection{Long-horizon Execution}
Despite its strong task understanding and memory capabilities, GPT-6 Astra exhibits limitations in sustained, consequence-aware action planning. While the model can identify semantically appropriate subgoals, it often fails to anticipate how the physical consequences of current actions affect the feasibility of subsequent operations. Figure~\ref{fig:long_horizon} (a) illustrates this issue in the LIBERO-Plus task of opening the top drawer and placing a bowl inside. Although the model successfully opens the drawer, the extended drawer obstructs access to the bowl required for the next step. This failure reveals a gap between semantic task planning and physical action planning: the model correctly identifies what should be done but does not sufficiently consider how executing one action constrains the next. Similar difficulties in anticipating downstream motion feasibility also emerge in the LIBERO-GeoMotion and real-world experiments discussed later.

Another challenge arises from the accumulation of execution errors and recovery costs. In the bottle-disposal task from RoboTwin 2.0 (Figure~\ref{fig:long_horizon} (b)) and the table-organization task from RoboDojo (Figure~\ref{fig:long_horizon} (c)), failures at intermediate subgoals interrupt task progress and require repeated adjustments. These recovery attempts consume additional environment steps, leaving fewer opportunities to complete the remaining subgoals. As task sequences become longer, local failures and their associated recovery costs accumulate, making successful completion increasingly difficult under a limited execution budget.

Together, these observations highlight two limitations in long-horizon manipulation: insufficient anticipation of the downstream physical consequences of actions and inefficient recovery from intermediate failures. Although GPT-6 Astra can reason about task objectives and execute individual steps, maintaining a physically feasible and coherent action sequence remains challenging.

\subsubsection{Precision}
Precise manipulation remains a significant limitation for GPT-6 Astra, particularly when tasks require fine-grained positional and rotational adjustments under strict geometric constraints. Although the model can often identify targets and approach them correctly, it struggles to translate approximate spatial understanding into precise physical execution. Figure~\ref{fig:precision} illustrates this gap through representative manipulation failures. In the plate-pushing task (a), the model successfully moves the plate close to the target region but fails to satisfy the required positional accuracy. More demanding insertion tasks expose similar difficulties. In the key-insertion task (b), the model repeatedly attempts to grasp the key and subsequently fails to align it with the keyhole, rotating the key before achieving proper insertion. Likewise, in the charger-plugging task (c), the robot approaches the socket but fails to achieve the precise alignment required for successful engagement.

These cases reveal a recurring gap between coarse spatial localization and fine-grained motion control. Reaching the vicinity of a target is often insufficient, as successful manipulation requires reasoning about object geometry, contact constraints, and small positional or angular deviations throughout execution.

Nevertheless, these limitations can be partially alleviated through additional sensory feedback. As discussed in Sec.~\ref{sec:tactile_experiments}, tactile information enables GPT-6 Astra to detect contact anomalies and adjust its actions accordingly. In the UniVTAC HDMI-insertion and peg-insertion tasks shown in Figure~\ref{fig:precision} (d,e), the model progressively corrects its position and approach orientation through repeated contact-based adjustments, eventually completing the insertions. These results suggest that tactile feedback can help bridge the gap between approximate spatial reasoning and precise physical interaction. However, the need for repeated corrections indicates that reliable fine-grained control remains challenging, even with additional contact information.

\begin{figure*}[!t]
    \includegraphics[width=\textwidth, trim=0mm 0mm 0mm 0mm]{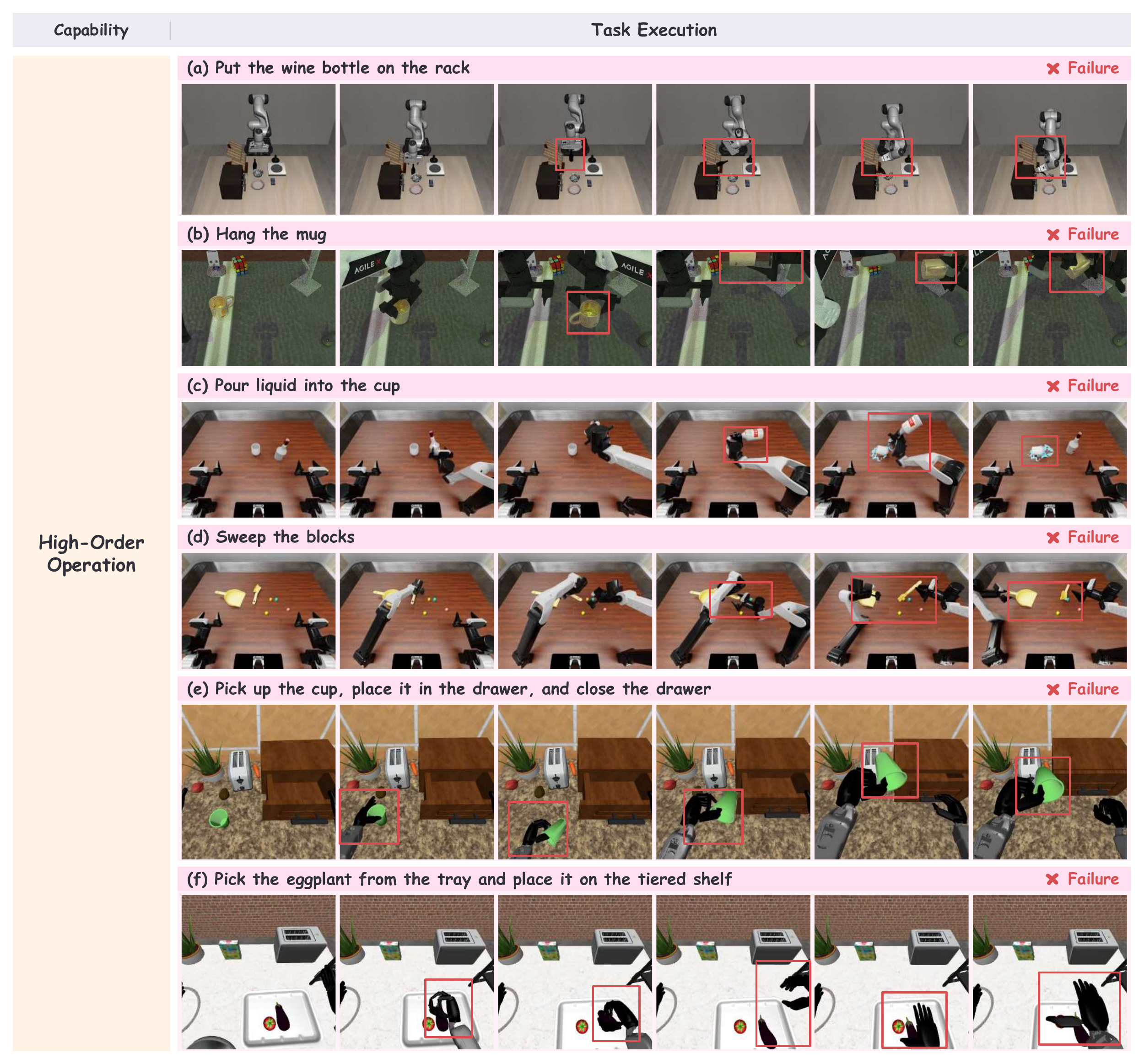}
    \caption{\textbf{The visualization of representative task executions illustrating limitations in \emph{high-order} operation for gripper-equipped robots and robots equipped with dexterous hands.} Green grounding boxes highlight successful actions, while red grounding boxes highlight failure-inducing actions.}
    \label{fig:high_order_operation}
\end{figure*}

\subsubsection{High-order Operation}
High-order manipulation remains a major challenge for GPT-6 Astra, particularly when tasks require coordinated control of object position, orientation, and contact throughout execution. Unlike simple pick-and-place operations, these tasks involve complex motion patterns such as rotation, pouring, and second-order actions, where manipulating one object affects another. Figure~\ref{fig:high_order_operation} illustrates these difficulties across several representative tasks.

In the wine-bottle placement task from LIBERO-Plus (a), the model attempts to rotate the bottle into an orientation suitable for the rack but fails to coordinate its motion, causing the bottle to fall. Similarly, in the mug-hanging task from RoboTwin 2.0 (b), the model lifts and rotates the mug but fails to align its handle with the rack. The pouring task from RoboDojo (c) further exposes difficulties in coordinating object orientation with ongoing interaction: although the model initiates the appropriate tilting motion, it fails to maintain the bottle opening above the receiving cup, causing liquid to spill outside. In the block-sweeping task (d), the model struggles to coordinate the broom's motion with the dustpan's position, leaving blocks scattered rather than successfully collecting them. These cases suggest that the primary difficulty lies not simply in selecting appropriate actions, but in maintaining the geometric and contact relationships required throughout their execution.

These limitations become more pronounced in RoboCasa GR1, where articulated hands introduce additional degrees of freedom and more complex contact interactions. In the cup-to-drawer task shown in Figure~\ref{fig:high_order_operation} (e), the model adopts an unsuitable grasp posture that complicates subsequent placement. Moreover, unintended hand contact pushes the previously opened drawer toward its closed position, reversing earlier progress. In the eggplant-placement task (f), repeated adjustments to the hand configuration fail to establish a stable grasp, preventing the robot from lifting the object. These failures reveal that the model struggles to coordinate arm motion, finger configurations, and object interactions, even when the semantic goal is straightforward.

Together, these observations highlight a fundamental limitation in continuous geometric and physical coordination. While GPT-6 Astra can identify appropriate manipulation strategies and initiate plausible motions, it struggles to maintain feasible object orientations and interaction constraints across successive actions. This gap becomes particularly apparent in rotation-dependent manipulation, motivating our subsequent investigation with LIBERO-GeoMotion.

\begin{figure*}
    \includegraphics[width=\textwidth]{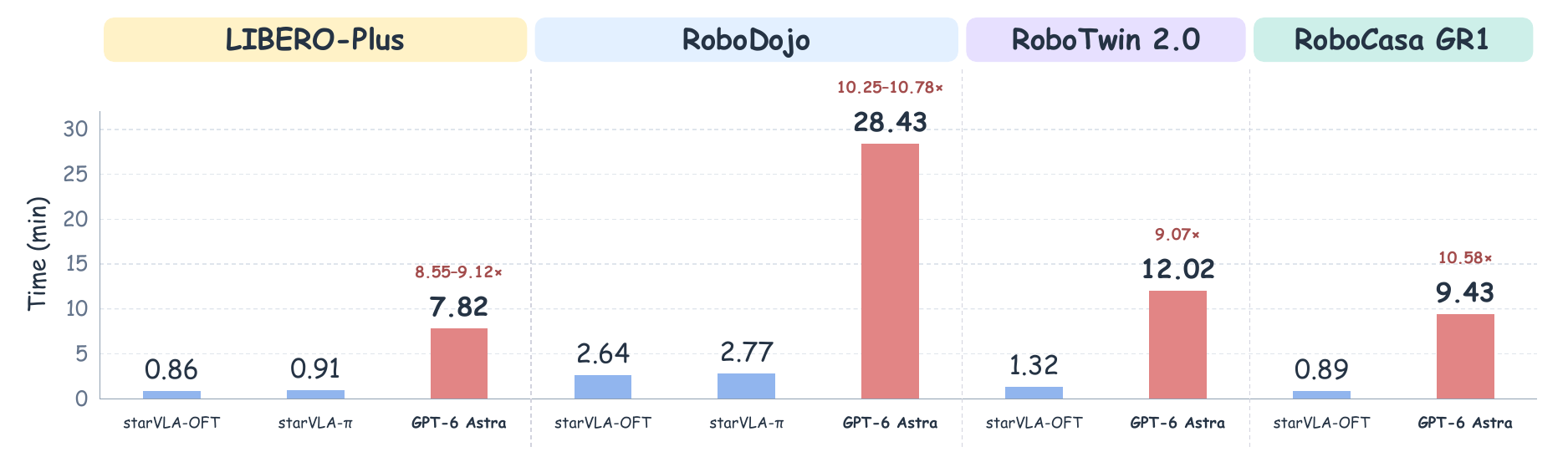}
    \caption{\textbf{The average execution latency for comparison.} Lower is better. Red bars denote GPT-6 Astra, and red annotations report its execution-time ratio relative to the other evaluated solutions.}
    \label{fig:runtime_comparison}
\end{figure*}

\subsubsection{Latency}

Despite its competitive zero-shot manipulation performance, GPT-6 Astra incurs substantial execution latency. As shown in Figure~\ref{fig:runtime_comparison}, its average task execution time on LIBERO-Plus reaches 7.82 minutes per episode, approximately 8.6--9.1 times that of the starVLA family~\cite{community2026starvla}. On RoboDojo, which exhibits the longest average execution time, GPT-6 Astra requires 28.43 minutes per episode, more than 10 times that of the starVLA family. A similar trend holds on RoboTwin 2.0 and RoboCasa GR1, where GPT-6 Astra takes approximately 9.1–10.6 times as long as starVLA-OFT. Although the evaluation settings are not directly comparable, these results suggest a substantial efficiency gap between general-purpose reasoning agents and specialized action policies.

This limitation also persists in real-world manipulation, where repeated action attempts and corrections can extend execution to approximately 20 minutes. Repeated multimodal reasoning, tool invocations, and iterative action correction introduce cumulative overhead throughout closed-loop execution. These observations highlight that, while general-purpose models enable flexible manipulation without task-specific policy training, their substantial inference and interaction costs remain a significant obstacle to efficient robotic deployment.

\section{LIBERO-GeoMotion}
\label{sec:libero_rotation}

Our systematic analysis reveals a recurring gap between GPT-6 Astra's spatial understanding and its ability to execute physically feasible manipulation motions. While the model performs well on straightforward pick-and-place tasks, it struggles with precise pose control, coordinated translation and rotation, and anticipating the physical consequences of actions. To further investigate these limitations, we introduce LIBERO-GeoMotion, a targeted benchmark designed to evaluate geometry-constrained manipulation while minimizing semantic complexity.

\begin{figure*}[!t]
    \centering
    \includegraphics[
        width=\textwidth,
        trim=0mm 10mm 0mm 10mm,
        clip
    ]{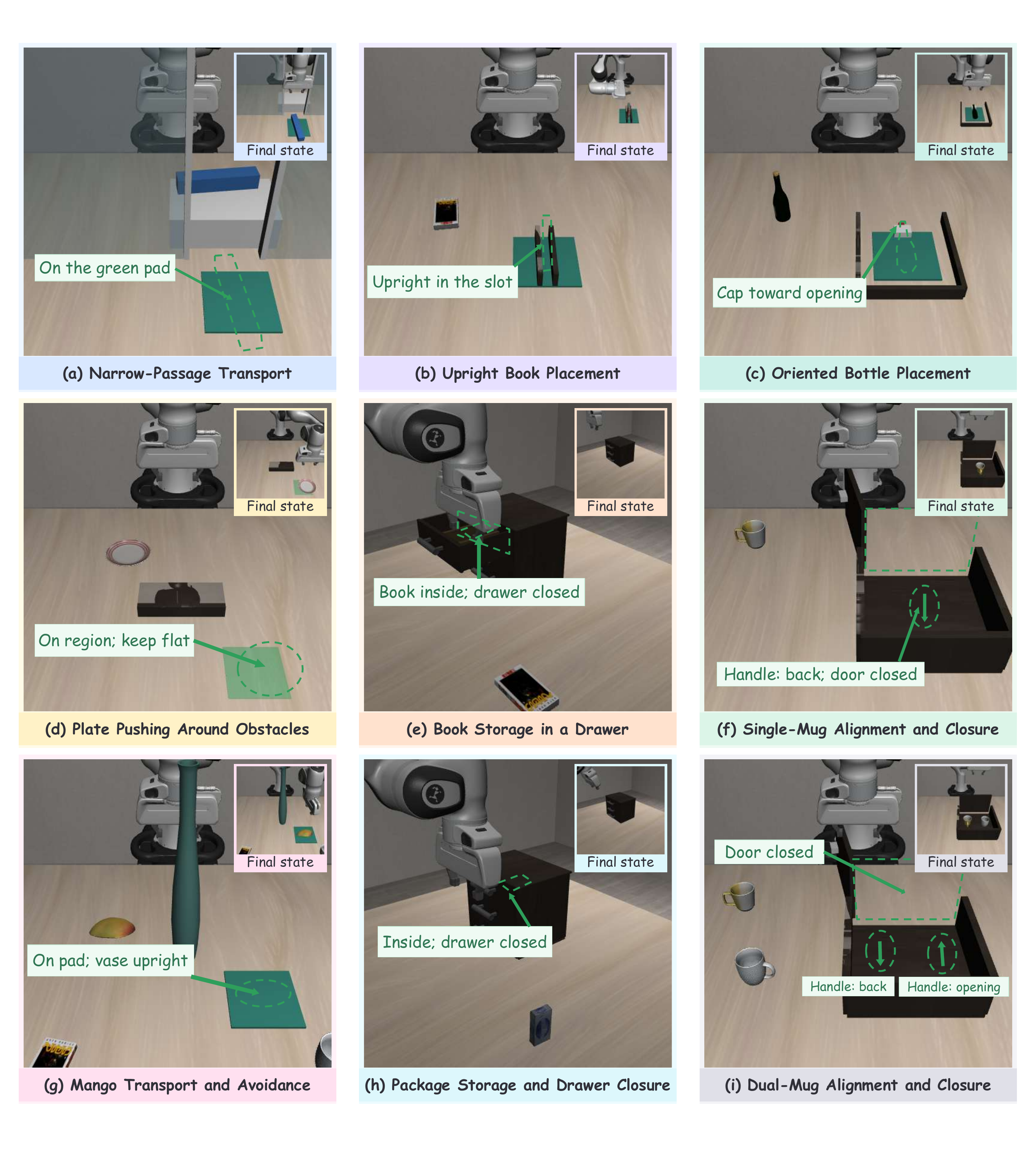}
    \caption{\textbf{Overview of the nine task families in LIBERO-GeoMotion.} Each panel illustrates a geometry-constrained manipulation task, with green dashed outlines indicating target poses or configurations and arrows highlighting spatial, orientation, or interaction requirements.}
    \label{fig:libero_rotation}
\end{figure*}

\subsection{Benchmark Design}

LIBERO-GeoMotion comprises 36 scene variants spanning nine task families, as illustrated in Figure~\ref{fig:libero_rotation}. The variants differ in initial object poses, target orientations, and spatial arrangements. By using familiar objects and relatively straightforward semantic goals, the benchmark places the primary demands on geometric reasoning and physical motion execution rather than complex task understanding.

The tasks cover three complementary manipulation requirements. First, \emph{grasp orientation alignment} requires selecting an appropriate end-effector orientation before grasping. For example, the mango task (g) requires rotating the gripper to grasp across the object's shorter axis while accounting for a nearby vase during subsequent transport.

Second, \emph{object reorientation for placement} requires satisfying explicit orientation constraints at the destination. Representative tasks include placing a book upright in a narrow slot (b), orienting a bottle within a rack with its cap facing the opening (c), and aligning mug handles toward specified directions (f, i). These operations require coordinating object orientation with the target configuration rather than simply reaching a desired position.

Third, \emph{coordinated motion under geometric and sequential constraints} requires maintaining feasible object poses throughout transport and interaction. The agent must reorient an elongated object to pass through a narrow opening (a), push a plate around obstacles while keeping it flat (d), or coordinate object reorientation and placement with drawer operations (e, h). The mug tasks additionally require closing the storage-box door after placement (f, i).

Across these tasks, instructions specify the desired outcomes without prescribing intermediate motions or trajectories. LIBERO-GeoMotion therefore provides a targeted evaluation of whether GPT-6 Astra can translate spatial understanding into precise and physically feasible manipulation sequences under geometric and sequential constraints.

\begin{table*}[t]
\centering
\caption{\textbf{Success rates (\%) on LIBERO-GeoMotion.} Columns correspond to the nine task families illustrated in Figure~\ref{fig:libero_rotation}. Each family contains four scene variants, and $\Delta$ denotes the improvement in percentage points.}
\label{tab:libero_rotation_icl}
\setlength{\tabcolsep}{4pt}
\renewcommand{\arraystretch}{1.15}
\begin{tabular}{lccccccccc}
\toprule
Task
& \shortstack{(a)\\Passage}
& \shortstack{(b)\\Book slot}
& \shortstack{(c)\\Bottle}
& \shortstack{(d)\\Plate}
& \shortstack{(e)\\Book drawer}
& \shortstack{(f)\\Mug}
& \shortstack{(g)\\Mango}
& \shortstack{(h)\\Package}
& \shortstack{(i)\\Two mugs} \\
\midrule
w/o ICL
& 0.0 & 0.0 & 50.0 & 0.0 & 0.0 & 0.0 & 50.0 & 0.0 & 0.0 \\
w/ ICL
& 25.0 & 50.0 & 75.0 & 25.0 & 25.0 & 75.0 & 75.0 & 0.0 & 100.0 \\
$\Delta$ (pp)
& +25.0 & +50.0 & +25.0 & +25.0 & +25.0 & +75.0 & +25.0 & 0.0 & +100.0 \\
\bottomrule
\end{tabular}
\end{table*}

\subsection{Overall Performance and ICL Adaptation}

Table~\ref{tab:libero_rotation_icl} summarizes GPT-6 Astra's performance under zero-shot and in-context learning (ICL) settings. Without contextual demonstrations, the model achieves an overall success rate of only 11.1\%, corresponding to 4 successful executions across 36 scene variants. Seven of the nine task families achieve zero success, with only bottle placement and mango manipulation reaching 50.0\%. This contrasts sharply with its competitive performance on several established manipulation benchmarks, revealing that proficiency in conventional manipulation tasks does not readily extend to tasks requiring complex geometric coordination and precise motion execution.

Providing contextual demonstrations substantially improves overall success to 50.0\%, with gains observed in eight of the nine task families. Particularly notable improvements appear in mug manipulation, where the single-mug task increases from 0\% to 75.0\% and the dual-mug task from 0\% to 100.0\%. These improvements suggest that contextual demonstrations provide useful motion references for selecting grasp orientations, adjusting object poses, and organizing manipulation sequences.

\begin{figure*}[!t]
    \centering
    \includegraphics[width=\textwidth]{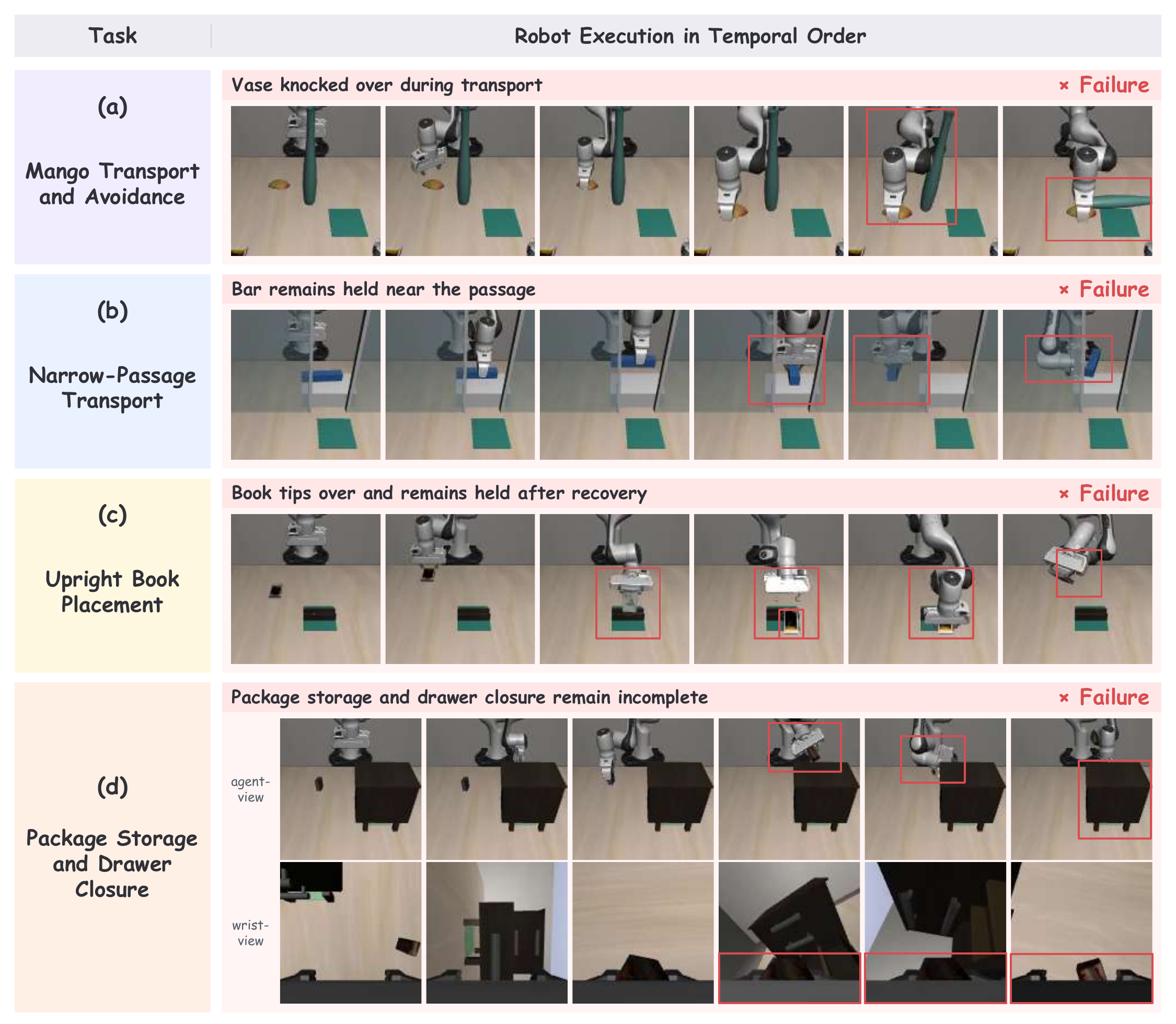}
    \caption{\textbf{Representative failure trajectories in LIBERO-GeoMotion.}
    (a) Mango transport: unintended contact knocks over a nearby vase.
    (b) Narrow-passage transport: unsuccessful coordination of translation and rotation prevents the elongated bar from passing through the opening.
    (c) Upright book placement: the book tips over during manipulation, and subsequent recovery attempts fail.
    (d) Package storage: the package remains outside the drawer, leaving placement and closure incomplete.
    Red boxes highlight representative failure-related states.}
    \label{fig:libero_rotation_failure}
\end{figure*}

Nevertheless, substantial limitations remain even with ICL. Several task families achieve success rates of only 25.0\%, while package storage remains entirely unsuccessful. These results indicate that contextual guidance can alleviate some geometric execution difficulties but does not consistently enable the coordinated motions required to satisfy complex physical constraints. To further examine these limitations, we analyze four representative failure trajectories in Figure~\ref{fig:libero_rotation_failure}.

\subsection{Performance Analysis}

\noindent\textbf{Insufficient Consequence-aware Action Planning.}
A recurring limitation is the model's difficulty in anticipating the physical consequences of planned motions. In the mango transport task shown in Figure~\ref{fig:libero_rotation_failure} (a), GPT-6 Astra attempts to transport the mango toward the target region but knocks over the nearby vase during execution. Although the model identifies the intended manipulation goal, it fails to adequately account for the space occupied by the moving object and its potential interaction with surrounding obstacles.

This failure reveals a limitation beyond inaccurate target localization or grasp selection. Successful transport requires anticipating how the object's geometry and movement may affect the surrounding environment throughout execution. In this case, the model generates a locally plausible motion without sufficiently considering its physical consequences. This behavior reinforces our earlier finding that GPT-6 Astra struggles with consequence-aware action planning, where selecting an appropriate immediate action does not guarantee a feasible or safe subsequent state.

\noindent\textbf{Coordinated Translation and Rotation under Constraints.}
A more direct limitation in complex motion execution appears in the narrow-passage transport task shown in Figure~\ref{fig:libero_rotation_failure} (b). GPT-6 Astra successfully approaches and grasps the elongated bar but fails to coordinate its orientation and translation while navigating the constrained passage. The bar becomes stuck near the opening, and the model cannot complete the required transport and placement.

Unlike conventional pick-and-place manipulation, this task requires continuous coordination of object position and orientation. The bar must be rotated into a feasible configuration while its translation is adjusted to satisfy the passage constraints. Simply predicting an appropriate final orientation is insufficient, as the object must remain geometrically feasible throughout the movement.

The observed failure highlights the model's difficulty in generating complex rotational motions coupled with translational control. Although GPT-6 Astra can initiate appropriate local movements, it does not consistently coordinate them into the continuous geometric transformations required for successful execution. This finding aligns with the limitations in high-order operations identified in our systematic analysis, where maintaining geometric relationships throughout an interaction is more challenging than selecting the intended operation.

\noindent\textbf{Precision and Recovery Limitations.}
Precise pose control remains another challenge in geometry-constrained manipulation. In the upright book-placement task shown in Figure~\ref{fig:libero_rotation_failure} (c), GPT-6 Astra attempts to orient the book for insertion into the narrow organizer. However, the book tips over during manipulation, and subsequent recovery attempts fail to restore a configuration suitable for placement. The model ultimately retains the book without completing the task.

This failure illustrates the difficulty of translating an understood target configuration into precise physical execution. Successful book placement requires accurate control of both position and orientation, particularly when the target slot imposes narrow geometric tolerances. Once the book deviates from the intended pose, the model must also identify an effective recovery strategy and restore a suitable configuration before attempting placement again.

The unsuccessful execution therefore reflects limitations in both fine-grained motion control and corrective manipulation. Consistent with our earlier precision analysis, GPT-6 Astra can reason about the desired spatial arrangement but struggles to achieve and maintain the exact physical configuration required for successful interaction.

\noindent\textbf{Challenges in Multi-step Motion Coordination.}
The package-storage task further exposes difficulties when object reorientation must be coordinated with multiple dependent interactions. As shown in Figure~\ref{fig:libero_rotation_failure} (d), GPT-6 Astra attempts to manipulate the package but fails to place it inside the drawer. The package remains on the tabletop, leaving both the storage and drawer-closure requirements incomplete.

Unlike a single-step placement task, package storage requires a sequence of coordinated operations, including grasping, reorienting the package, placing it within the available storage space, and closing the drawer. These operations impose interdependent geometric constraints: the object must be manipulated into a suitable orientation and position before subsequent interactions can be completed.

The observed failure is consistent with the difficulties in sustained action planning and complex motion coordination identified in our systematic analysis. Notably, package storage remains unsuccessful across all four evaluated variants even with ICL. This suggests that contextual motion references alone are insufficient to ensure successful execution when multiple geometric and sequential requirements must be satisfied.

\subsection{Discussion}

Overall, LIBERO-GeoMotion further substantiates the capability limitations identified in our systematic analysis through targeted geometry-constrained manipulation tasks. The representative failures reveal several distinct but related challenges: insufficient anticipation of physical action consequences during mango transport, difficulties coordinating translation and rotation in narrow-passage manipulation, inaccurate pose control and recovery in book placement, and unreliable execution of sequential interactions in package storage.

These findings suggest that GPT-6 Astra's manipulation limitations extend beyond isolated errors in spatial perception or target-pose estimation. Although the model can understand desired object configurations and initiate plausible actions, it does not consistently generate the precise, coordinated, and physically feasible motion sequences required to realize them. In particular, complex rotational control remains a major bottleneck when orientation changes must be coordinated with translation, contact, and subsequent actions.

While embodied ICL substantially improves performance, the remaining failures indicate that contextual demonstrations cannot fully compensate for these limitations. Together, our findings highlight a persistent gap between spatial reasoning and reliable geometric motion execution, providing further evidence of the physical control challenges that general-purpose models must overcome for complex robotic manipulation.

\section{Real-Robot Experiments}
\label{sec:real_world}

\subsection{Experimental setup}
To evaluate GPT-6 Astra's performance in physical environments, we conduct real-world experiments on a ROKAE Helios-series robot\footnote{\href{https://www.rokae.com/en/product/show/596/Wheeled-Dual-Arm-Robot-Helios.html}{ROKAE Helios official product page.}}. The robot comprises a wheeled mobile base and a torso carrying two 7-degree-of-freedom (DoF) arms, each with force-control capabilities and a parallel-jaw gripper. We design five manipulation tasks that probe basic object manipulation, grasp-pose selection under geometric constraints, precise placement, manipulation strategy selection, and open-world semantic rearrangement.

\begin{figure*}[h]
    \centering
    \includegraphics[width=\textwidth]{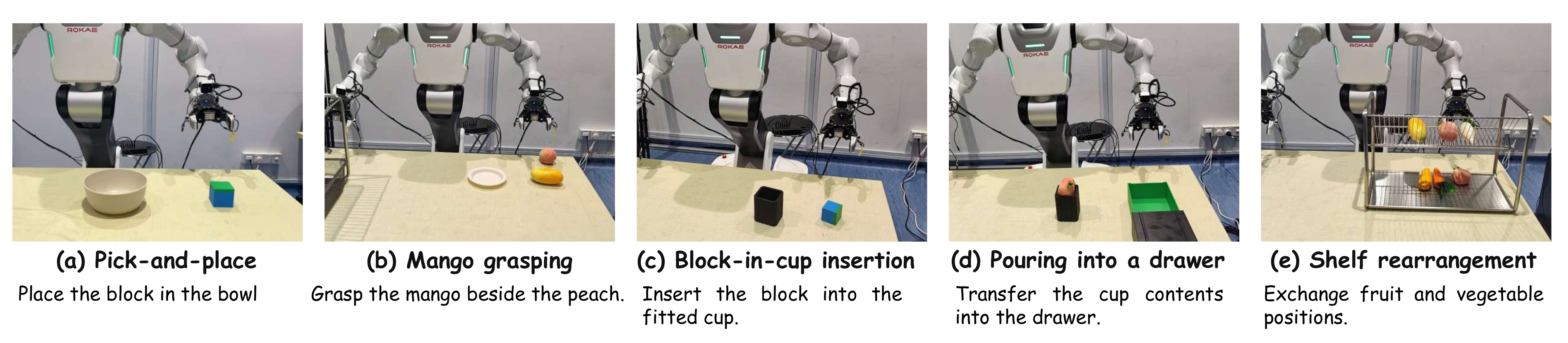}
\end{figure*}

\begin{enumerate}
    \item \textbf{Pick-and-place.}
    The robot picks up a small block and places it on a plate. This task serves as a basic evaluation of object localization, grasping, and placement.

    \item \textbf{Constrained mango grasping.}
    The robot must grasp a mango whose length along its major axis exceeds the maximum gripper opening. A nearby peach acts as an obstacle, restricting downward motion and preventing the gripper from reaching the lower part of the mango. These constraints require the robot to select a feasible grasp orientation while accounting for both object geometry and the surrounding obstacle.

    \item \textbf{Block-in-cup insertion.}
    The robot grasps a block and inserts it into a custom 3D-printed cup. For safe and stable manipulation, the gripper must align with opposing faces of the block rather than grasp its edges. The cup closely matches the block's dimensions, leaving little clearance and requiring precise adjustment of both position and orientation during insertion.

    \item \textbf{Pouring into a drawer.}
    The robot transfers all objects from a cup into a drawer. The objects inside the cup are inaccessible to direct grasping, requiring the robot to grasp and tilt the cup to pour out its contents. This task evaluates whether the robot can select an appropriate manipulation strategy and control the cup's orientation to complete the transfer.

    \item \textbf{Shelf rearrangement.}
    The robot exchanges the positions of fruits and vegetables on a shelf. We use this as an open-world manipulation task involving multiple objects and semantic categories, requiring the robot to interpret the rearrangement goal, identify relevant objects, and sequence the necessary pick-and-place actions.
\end{enumerate}

\begin{table}[t]
    \centering
    \small
    \setlength{\tabcolsep}{4pt}
    \caption{Real-world success rates under zero-shot and
    one-example ICL conditions. Each task is evaluated over
    five trials per condition. Counts indicate successful
    trials divided by total trials.}
    \label{tab:real_world}
    \begin{tabular}{lcc}
        \toprule
        Task & Zero-shot & One-example ICL \\
        \midrule
        Pick-and-place
            & 100\% (5/5) & 100\% (5/5) \\
        Mango grasping
            & 0\% (0/5)   & 60\% (3/5) \\
        Block-in-cup insertion
            & 40\% (2/5)  & 80\% (4/5) \\
        Pouring into a drawer
            & 40\% (2/5)  & 100\% (5/5) \\
        Shelf rearrangement
            & 20\% (1/5)  & 60\% (3/5) \\
        \midrule
        Overall
            & 40\% (10/25) & 80\% (20/25) \\
        \bottomrule
    \end{tabular}
\end{table}

\subsection{Evaluation protocol}
We evaluate each task under two conditions: \emph{zero-shot}, where GPT-6 Astra receives no demonstration, and \emph{one-example in-context learning (ICL)}, where the model is provided with a single demonstration of a similar task. We conduct five trials for each task under each condition, resulting in 50 trials in total. We report task success rates, computed as the fraction of trials in which the specified task goal is achieved.

\subsection{Results}
As shown in Table~\ref{tab:real_world}, GPT-6 Astra achieves an overall success rate of 40\% in the zero-shot setting and 80\% with one-example ICL. Basic pick-and-place succeeds in all five trials under both conditions, while ICL improves the observed success rate on each of the other four tasks. The largest gains occur in constrained mango grasping and pouring into a drawer, both improving by 60 percentage points. These results suggest that a single related demonstration can help it address geometric constraints and select effective manipulation strategies in real-world tasks.
\subsection{Real-World Performance Analysis}
\label{sec:real_world_analysis}



 \begin{figure*}
    \centering
    \includegraphics[width=\textwidth]{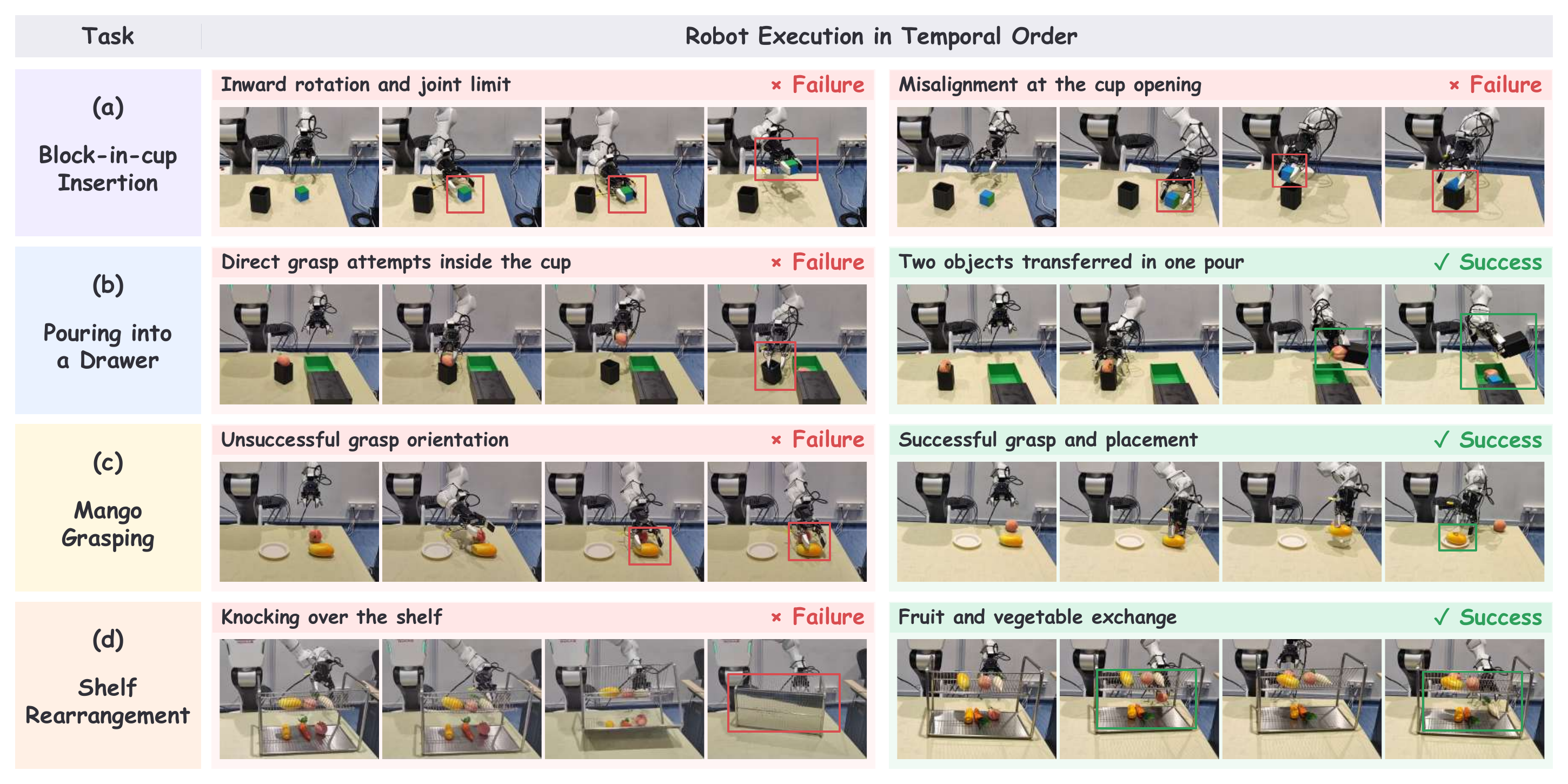}
    \caption{\textbf{Representative real-world execution trajectories of GPT-6 Astra.} Green grounding boxes highlight successful actions, while red grounding boxes highlight failure-inducing actions.}
    \label{fig:real_world_visualization}
\end{figure*}

\label{sec:real_world_results}

We further evaluate GPT-6 Astra on five real-world manipulation tasks under zero-shot and one-example in-context learning (ICL) settings, with five trials per task. Despite its promising performance on several simulation benchmarks, GPT-6 Astra achieves only 40.0\% overall success in real-world zero-shot execution. While basic pick-and-place succeeds in all five trials, performance drops substantially on more demanding tasks. Block-in-cup insertion and pouring achieve only 40.0\% success, shelf rearrangement reaches 20.0\%, and constrained mango grasping fails in all five trials. These results indicate that although certain basic manipulation capabilities can be directly deployed without task-specific training, reliable execution remains challenging under more complex physical and geometric constraints.

In contrast, providing a single contextual demonstration increases overall success from 40.0\% to 80.0\%, with improvements across all four challenging tasks. This substantial gain highlights the potential of embodied ICL to improve real-world manipulation without parameter updates. Nevertheless, the remaining failures reveal persistent limitations in motion planning, geometric precision, and physical interaction, which we examine below.

\noindent\textbf{Rotation and Consequence-aware Motion Planning.}
The real-world experiments reinforce the limitations in rotational control and consequence-aware action planning identified in our systematic analysis and LIBERO-GeoMotion. In the block-in-cup task shown in Figure~\ref{fig:real_world_visualization} (a), GPT-6 Astra can establish an appropriate grasp but subsequently rotates its wrist inward, reaching a joint limit that prevents the required placement. Although the initial grasp is successful, the selected wrist configuration does not preserve sufficient motion range for the next operation. This reveals a failure to anticipate how an immediate action constrains subsequent physical execution.

A similar limitation appears in the mango-grasping task shown in Figure~\ref{fig:real_world_visualization} (c), which extends the rotation-dependent grasping problem examined in LIBERO-GeoMotion to a more constrained physical environment. The mango's major axis exceeds the gripper opening, requiring the model to select a suitable grasp orientation across its shorter axis. However, GPT-6 Astra sometimes fails to perform the necessary reorientation and instead attempts an infeasible grasp. In other attempts, it lowers the gripper before establishing a collision-free approach, causing the arm to encounter the nearby peach. A feasible strategy requires coordinating grasp orientation with an oblique approach that avoids the obstacle.

These cases expose a common limitation: GPT-6 Astra can propose locally plausible actions but does not consistently anticipate their downstream kinematic and geometric consequences. The difficulty lies not merely in determining the desired object pose, but in planning a sequence of physically feasible motions that preserves the ability to complete subsequent operations.

\noindent\textbf{Precision under Tight Geometric Constraints.}
Precise manipulation remains another significant challenge in real-world execution. The block-in-cup task requires inserting a block into a closely fitting container, leaving little tolerance for positional and angular errors. As shown in the right-hand sequence of Figure~\ref{fig:real_world_visualization} (a), even when the model successfully grasps and transports the block toward the opening, residual pose misalignment prevents successful insertion.

This failure is consistent with the precision limitations observed in simulation. GPT-6 Astra can often identify the correct target and approximate the required motion, yet struggles to satisfy the fine-grained geometric constraints needed for physical engagement. Although ICL improves block-in-cup success from 40.0\% to 80.0\%, accurate pose alignment remains a challenge in the remaining unsuccessful executions.

\noindent\textbf{Safety Limitations of Direct Pose-based Control.}
Beyond task success, the experiments reveal potential safety concerns when directly translating model-generated end-effector pose increments into robot movements. In the shelf-rearrangement task shown in Figure~\ref{fig:real_world_visualization} (d), GPT-6 Astra attempts to manipulate objects on the shelf but unintentionally knocks over the shelf during execution. This failure demonstrates that an action directed toward a semantically correct goal can still produce undesirable physical consequences through unintended contact with the environment.

The joint-limit issue in block insertion and obstacle interference in mango grasping further illustrate the importance of motion feasibility. Although low-level controllers can enforce kinematic and joint-limit constraints, these checks do not guarantee collision-free trajectories or safe object interactions. Directly generating end-effector pose changes without sufficiently considering the robot's configuration, surrounding geometry, and future motion requirements therefore remains insufficient for reliable physical execution. While these cases do not constitute a comprehensive safety evaluation, they expose practical risks that must be addressed before deploying general-purpose models as autonomous robotic controllers.

\noindent\textbf{Embodied ICL Enables Action-level Adaptation.}
The real-world results further demonstrate that embodied ICL can improve not only semantic task understanding but also manipulation strategies and motor execution. Quantitatively, contextual demonstrations improve constrained mango grasping from 0.0\% to 60.0\%, block-in-cup insertion from 40.0\% to 80.0\%, pouring from 40.0\% to 100.0\%, and shelf rearrangement from 20.0\% to 60.0\%. These improvements are particularly notable because the tasks involve different physical requirements, including grasp orientation, precise alignment, object reorientation, and multi-step manipulation.

The execution trajectories in Figure~\ref{fig:real_world_visualization} provide further insight into these improvements. In the mango-grasping task (c), contextual demonstrations help the model select a more appropriate grasp orientation and approach direction, overcoming some of the geometric constraints that prevent successful zero-shot execution. Similarly, demonstrations provide useful rotation and grasping references for block-in-cup insertion, allowing the model to adopt more feasible manipulation motions.

The pouring task (b) provides particularly interesting evidence of action-level adaptation. Without demonstrations, GPT-6 Astra sometimes repeatedly attempts to grasp inaccessible objects inside the cup. With ICL, the model more consistently selects and executes the appropriate pouring strategy. Notably, one successful execution transfers two objects into the drawer in a single pour, although this specific operation was not demonstrated. This suggests that the model can adapt a learned manipulation strategy to a new physical configuration rather than simply reproducing the demonstrated action sequence.

Together, the quantitative improvements and execution behaviors provide evidence that embodied ICL extends beyond semantic guidance to the adaptation of grasping strategies, rotation patterns, and manipulation procedures. However, the remaining failures, particularly in constrained grasping, precise insertion, and long-sequence execution, indicate that contextual demonstrations cannot fully compensate for limitations in physical planning and control.

\subsection{Discussion}
Overall, our real-world experiments further corroborate the capability boundaries identified in our systematic analysis and LIBERO-GeoMotion. Although GPT-6 Astra demonstrates the potential for zero-shot deployment on basic manipulation tasks, its limited overall success and recurring failures in constrained environments reveal substantial weaknesses in consequence-aware action planning, fine-grained precision, and continuous motion coordination. Unintended contacts, including knocking over surrounding objects and reaching joint limits, further expose safety concerns associated with directly generated robot actions.

Meanwhile, the substantial improvements achieved through ICL reinforce our earlier finding that general-purpose models can acquire and adapt simple manipulation strategies from contextual demonstrations without additional training. Nevertheless, even with such adaptation, robust physical execution is not guaranteed. These findings highlight a fundamental distinction between the emerging ability of general-purpose models to perform robotic manipulation and the safety, efficiency, and reliability required for practical real-world deployment.

\section{Conclusion}
In this work, we presented a systematic evaluation of GPT-6 Astra for direct closed-loop robotic manipulation through GPT-GEAR, a lightweight agent framework that requires neither specialized action policies nor task-specific parameter updates. Spanning five established benchmarks, six capability dimensions, our new LIBERO-GeoMotion benchmark, and real-robot experiments, the study provides the community with a broad empirical baseline for general-purpose models as robotic controllers. GPT-6 Astra exhibits \textbf{emerging embodied capabilities}: it exploits depth information, adapts manipulation strategies from contextual demonstrations, and uses tactile feedback for corrective actions—capabilities whose effectiveness depends on how sensory evidence and demonstrations are presented, and which support competitive benchmark results. However, we identify a \textbf{persistent gap between semantic reasoning and physical execution}: precise alignment, coordinated motion, consequence-aware planning, and safety remain unreliable, and while embodied in-context learning enables adaptation of strategies and motions, contextual guidance does not fully resolve these deficits, with substantial inference and interaction overhead further impeding deployment. The strengths of general-purpose models in task understanding, generalization, and memory point toward a promising pathway for general-purpose robotic control; the remaining challenges in translating flexible reasoning into safe, efficient, and reliable action delineate where future advances are needed.

\section{Acknowledgements}
This work is funded by Fundamental and Interdisciplinary Disciplines Breakthrough Plan of the Ministry of Education of China (JYB2025XDXM902) and National Natural Science Foundation of China (Grant No. 62606219).

\IfFileExists{reference.bib}{%
  \bibliographystyle{plainnat}
  \bibliography{reference}
}{%
  \typeout{WARNING: reference.bib is missing; bibliography omitted.}
}

\end{document}